\documentclass[sn-mathphys,Numbered]{sn-jnl}% Math and 
\usepackage{graphicx}%
\usepackage{multirow}%
\usepackage{amsmath,amssymb,amsfonts}%
\usepackage{amsthm}%
\usepackage{mathrsfs}%
\usepackage[title]{appendix}%
\usepackage{xcolor}%
\usepackage{textcomp}%
\usepackage{manyfoot}%
\usepackage{booktabs}%
\usepackage{subcaption}
\usepackage{algorithm}%
\usepackage{algorithmicx}%
\usepackage{algpseudocode}%
\usepackage{listings}%
\usepackage{lipsum}
\usepackage{ulem}
\usepackage{wrapfig}
\usepackage{enumitem}
\usepackage{indentfirst}
\usepackage{lineno}
\begin{document}

\title[Article Title]{Foundation Models for Generalist Geospatial Artificial Intelligence}

\author{\fnm{Johannes} \sur{Jakubik}\textsuperscript{1,$\ddagger$}}
\author{\fnm{Sujit} \sur{Roy}\textsuperscript{3,$\dagger, \ddagger$}}
\author{\fnm{C. E.} \sur{Phillips}\textsuperscript{3,$\dagger$}}
\author{\fnm{Paolo} \sur{Fraccaro}\textsuperscript{1,$\dagger$}}
\author{\fnm{Denys} \sur{Godwin}\textsuperscript{4}}
\author{\fnm{Bianca} \sur{Zadrozny}\textsuperscript{1}}
\author{\fnm{Daniela} \sur{Szwarcman}\textsuperscript{1}}
\author{\fnm{Carlos} \sur{Gomes}\textsuperscript{1}}
\author{\fnm{Gabby} \sur{Nyirjesy}\textsuperscript{1}}
\author{\fnm{Blair} \sur{Edwards}\textsuperscript{1}}
\author{\fnm{Daiki} \sur{Kimura}\textsuperscript{1}}
\author{\fnm{Naomi} \sur{Simumba}\textsuperscript{1}}
\author{\fnm{Linsong} \sur{Chu}\textsuperscript{1}}
\author{\fnm{S. Karthik} \sur{Mukkavilli}\textsuperscript{1}}
\author{\fnm{Devyani} \sur{Lambhate}\textsuperscript{1}}
\author{\fnm{Kamal} \sur{Das}\textsuperscript{1}}
\author{\fnm{Ranjini} \sur{Bangalore}\textsuperscript{1}}
\author{\fnm{Dario} \sur{Oliveira}\textsuperscript{1}}
\author{\fnm{Michal} \sur{Muszynski}\textsuperscript{1}}
\author{\fnm{Kumar} \sur{Ankur}\textsuperscript{3}}
\author{\fnm{Muthukumaran} \sur{Ramasubramanian}\textsuperscript{3}}
\author{\fnm{Iksha} \sur{Gurung}\textsuperscript{3}}

\author{\fnm{Sam} \sur{Khallaghi}\textsuperscript{4}}
\author{\fnm{Hanxi (Steve)} \sur{Li}\textsuperscript{4}}
\author{\fnm{Michael} \sur{Cecil}\textsuperscript{4}}

\author{\fnm{Maryam} \sur{Ahmadi}\textsuperscript{4}}
\author{\fnm{Fatemeh} \sur{Kordi}\textsuperscript{4}}
\author{\fnm{Hamed} \sur{Alemohammad}\textsuperscript{4,5}}

\author{\fnm{Manil} \sur{Maskey}\textsuperscript{2}}
\author{\fnm{Raghu} \sur{Ganti}\textsuperscript{1}}
\author{\fnm{Kommy} \sur{Weldemariam}\textsuperscript{1,$\ddagger$}}
\author{\fnm{Rahul} \sur{Ramachandran}\textsuperscript{2,$\ddagger$}}

\affil[1]{\orgname{IBM Research}}

\affil[2]{\orgname{NASA Marshall Space Flight Center}, \city{Huntsville}, \state{AL}, \country{USA}}

\affil[3]{\orgdiv{Earth System Science Center}, \orgname{The University of Alabama in Huntsville}, \state{AL}, \country{USA}}

\affil[4]{\orgdiv{Center for Geospatial Analytics}, \orgname{Clark University}, \city{Worcester}, \state{MA}, \country{USA}}

\affil[5]{\orgdiv{Graduate School of Geography}, \orgname{Clark University}, \city{Worcester}, \state{MA}, \country{USA}}

\makeatletter\def\Hy@Warning#1{}\makeatother
\Footnotetext{$\dagger$}{Equal contribution. $\ddagger$ Corresponding authors: johannes.jakubik@ibm.com, sujit.roy@nasa.gov, kommy@ibm.com, rahul.ramachandran@nasa.gov.}

\abstract{Significant progress in the development of highly adaptable and reusable Artificial Intelligence (AI) models is expected to have a significant impact on Earth science and remote sensing. Foundation models are pre-trained on large unlabeled datasets through self-supervision, and then fine-tuned for various downstream tasks with small labeled datasets. There is an increasing interest within the scientific community to investigate whether this approach can be successfully applied to domains beyond natural language processing to effectively build generalist AI models that exploit multi-sensor data. This paper introduces a first-of-its-kind framework for the efficient pre-training and fine-tuning of foundational models on extensive geospatial data. We have utilized this framework to create Prithvi, a transformer-based geospatial foundational model pre-trained on more than 1TB of multispectral satellite imagery from the Harmonized Landsat-Sentinel 2 (HLS) dataset. Our study demonstrates the efficacy of our framework in successfully fine-tuning Prithvi to a range of Earth observation tasks that have not been tackled by previous work on foundation models involving multi-temporal cloud gap imputation, flood mapping, wildfire scar segmentation, and multi-temporal crop segmentation. We thoroughly examine and assess the effect of Prithvi's pre-trained weights on downstream tasks. We compare learning curves between 1) fine-tuning the entire model, 2) fine-tuning solely the decoder for the downstream task, and 3) training the model without utilizing Prithvi's pre-trained weights. Our experiments show that the pre-trained model accelerates the fine-tuning process compared to leveraging randomly initialized weights. In addition, pre-trained Prithvi compares well against the state-of-the-art on downstream tasks, e.g., outperforming a conditional GAN model in multi-temporal cloud imputation by up to 5pp (or 5.7\%) in the structural similarity index. Finally, due to the limited availability of labeled data in the field of Earth observation, we gradually reduce the quantity of available labeled data for refining the model to evaluate data efficiency and demonstrate that data can be decreased significantly without affecting the model's accuracy. The pre-trained 100 million parameter model and corresponding fine-tuning workflows have been released publicly as open source contributions to the global Earth sciences community through Hugging Face\footnotemark.
} 

\Footnotetext{$\ast$}{\url{https://huggingface.co/ibm-nasa-geospatial}}

\maketitle

\section{Introduction}\label{sec1}
As data availability and artificial intelligence (AI) model size continue to increase, developing and training models has become an ever-more costly endeavor. This is of particular relevance in the geosciences, which are observation-driven. In today's environment of vast remote sensing data volumes, the challenges of exploring and processing unlabeled data pose substantial obstacles to research endeavors. Therefore, machine learning and deep learning techniques are increasingly used to more efficiently analyze these datasets. However, the vast majority of AI models for geoscience and remote sensing still require labeled data for supervised approaches, which can be expensive to produce \cite{Hamed2020Advancing}. Further, these task-specific models do not generalize well in space and time, necessitating the construction of a new model for each application \cite{Hamed2020Advancing}. 
Foundation models have emerged to address these issues by leveraging self-supervision to pre-train on large unlabeled datasets. The foundation model can then be fine-tuned to a variety of downstream tasks using smaller labeled datasets. This task-agnostic paradigm has recently unlocked new levels of performance and emergent behaviors in domains such as natural language processing.
Thus, there has been a rising interest in the scientific community to better understand how this approach can be utilized in the geoscience and remote sensing domains to effectively build generalist models that can perform on a variety of tasks.

In this paper, we propose a first-of-its-kind framework for the creation of geospatial foundation models to accelerate the development and deployment of climate and sustainability applications. Our framework provides a distributed and scalable infrastructure for training, fine-tuning, and inference that connects directly to geospatial data sources via intelligent data discovery operators (e.g., sampling and pre-processing). We provide an overview of our framework in Figure \ref{fig:framework}. 
We used this framework to train Prithvi\footnote{Prithvi is the Sanskrit name for the Earth.}, a geospatial foundation model using multispectral satellite measurements from NASA's Harmonized Landsat Sentinel-2 (HLS). The pre-trained model is fine-tuned for various downstream applications such as multi-temporal cloud imputation, flood mapping, fire-scar segmentation, and multi-temporal crop segmentation. 
We comprehensively study and evaluate the performance of the model by examining a set of key research questions (RQs) considering development, impact, and collaboration in the field of AI for geoscience:

\begin{figure}[t]
    \centering
    \includegraphics[width=\textwidth]{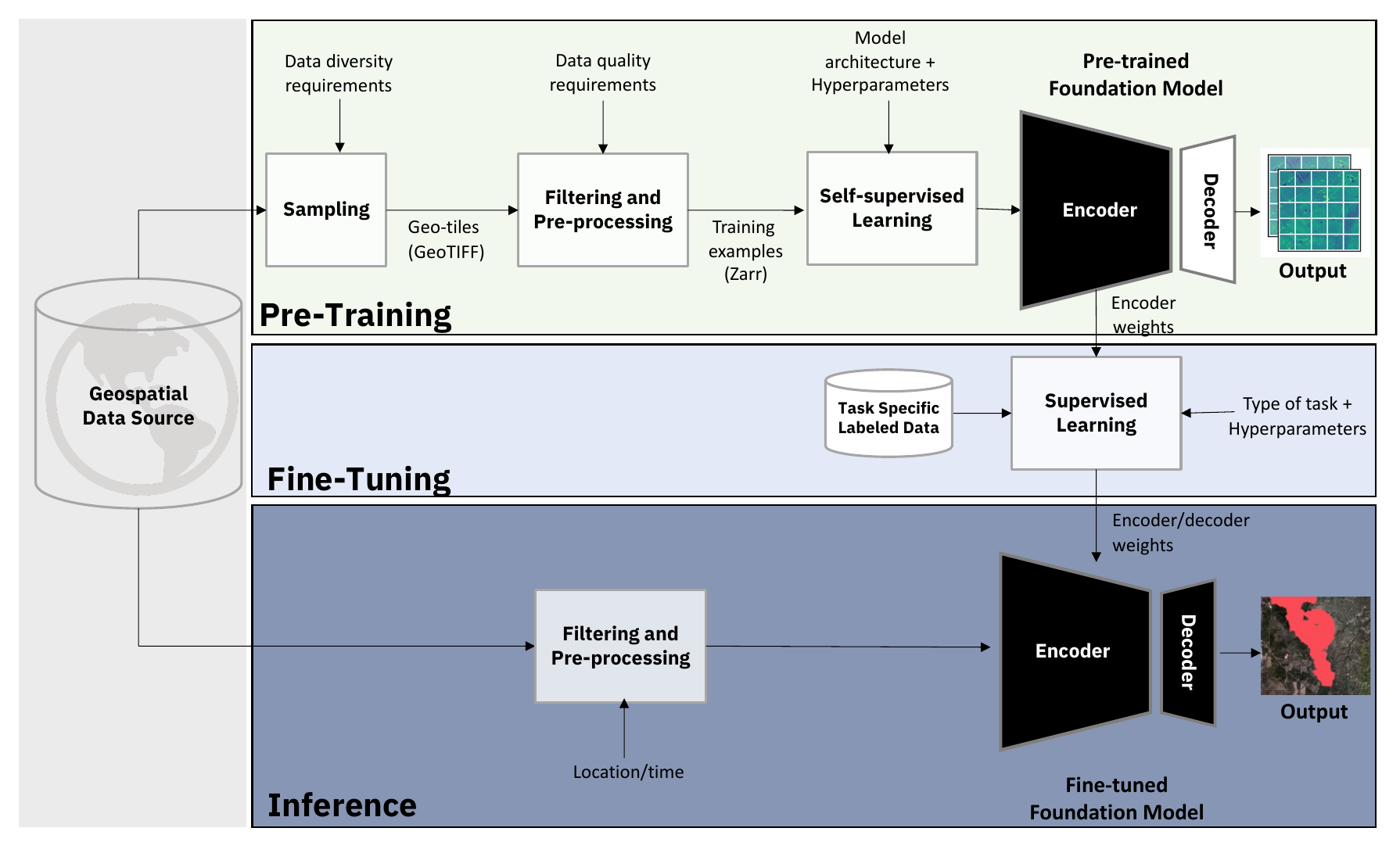}
    \caption{We propose a first-of-its-kind framework for the development of geospatial foundation models from raw satellite imagery, which we leverage to generate the Prithvi-100M model. The framework encompasses (1)~the sampling, filtering, and pre-processing of raw geospatial data and the self-supervised foundation model pretraining, (2)~the fine-tuning to specific downstream applications, and (3)~the inference process. 
    }
    \label{fig:framework}
\end{figure}

\newlist{myitemize}{itemize}{1}
\setlist[myitemize,1]{label={}, left=2em, labelsep=0em, itemindent=0em, align=left}

\begin{myitemize}
   \item
   \item RQ1: What are factors that play a key role in designing and evaluating foundation models in geoscience?
   \item
   \item RQ2: Given a large volume of remote sensing data, which may contain repetitive information across different ecosystems and landscapes, as well as noise inherent in the data acquisition process (i.e., from cloud occlusion or sensor malfunctioning), how can we efficiently pre-train a foundation model, ensuring noise removal and avoidance of redundancies?
   \item
   \item RQ3: Can foundation models exploit different features from the training data and generalize across different geoscience domains while needing significantly fewer labeled data to meet benchmark performance?
   \item
\end{myitemize}

In the remainder of this work, we comprehensively answer these key research questions for AI in geoscience in the era of large-scale self-supervised learning and foundation models. By sharing our insights and providing open-source access to our model architecture, pre-training weights, and inference service on HuggingFace, we support the acceleration of AI for geoscience.

\section{Background}

Foundation models are generalist AI models that are pre-trained on large unlabeled datasets through self-supervision and then fine-tuned for different downstream tasks. In recent years, they have been shown to be a very effective approach for natural language processing \cite{10.48550/arxiv.2302.13971} and computer vision tasks \cite{10.48550/arxiv.2111.11432}. Recent efforts often combine image and language into multi-modal foundation models (e.g., \cite{10.48550/arxiv.2306.11029,10.48550/arxiv.2205.01917,10.48550/arxiv.2209.07526}). This has led to an increasing interest in the scientific community to investigate whether this approach could be successfully applied to other domains to effectively build generalist AI models that make use of different types of data, beyond text and images, like \cite{satmae,10.48550/arxiv.1711.10398,10.1109/tgrs.2022.3194732,10.48550/arxiv.2304.05215,nguyen2023climax, mukkavilli2023ai}. For a more detailed perspective on weather and climate foundation models, we point the reader to our review in \cite{mukkavilli2023ai}. While many deep-learning frameworks have been applied to Earth Science tasks, we only review the most relevant comparisons to our presented framework \cite{ma2019deep, aleissaee2023transformers}. The first of these is the U-Net, a convolutional neural network proposed by Ronneberger et al. \cite{ronneberger2015u} in 2015. The U-Net is characterized by a two-leg design where the first leg is for feature extraction, and the second is the upper sampling portion \cite{10.1155/2022/1603273}. At each level, the sampling leg fuses the output of the prior layer with the adjacent layer of the feature extractor. This architecture has been deployed successfully for segmentation tasks, including buildings \cite{10.1155/2022/1603273} and waterbodies \cite{10.1109/lgrs.2020.3047918}. These tasks have been performed with both visible and hyperspectral imagery.

The Masked Autoencoder (MAE) \cite{he2022masked} is designed to reconstruct original signals from partial observations. MAE divides images into non-overlapping patches, randomly sampling some and masking others to reduce redundancy. Its encoder, based on the Vision Transformer (ViT) \cite{dosovitskiy2020image} model, processes only visible patches, optimizing computational efficiency. The decoder reconstructs images using encoded visible patches and mask tokens, primarily aiding in pre-training. The reconstruction process predicts pixel values for masked patches, with the difference between original and reconstructed images measured using Mean Squared Error (MSE). Efficiently designed, MAE's pre-training avoids specialized operations, making it a novel approach to image processing.

In the case of Earth sciences, there has been some work showing that variations of self-supervised learning schemes originally developed for computer vision could be adapted for pre-training models based on multi-spectral and temporal satellite imagery \cite{satmae}. There have also been recent proposals of foundation models for remote sensing \cite{10.48550/arxiv.1711.10398,10.1109/tgrs.2022.3194732,jakubik2023toward} with a billion-parameter model being demonstrated \cite{10.48550/arxiv.2304.05215}. These models, however, focus on aerial images from benchmark datasets that target object detection tasks and are limited to the visible bands (red, green, and blue). Thus, these models disregard the geospatial, multispectral, and temporal nature of satellite imagery, making them unsuitable for large scale Earth observation tasks. A first approach towards including these kinds of data in a self-supervised pre-training is Presto \cite{tseng2023lightweight}. Presto has been developed to facilitate lightweight computations and, therefore, is comparatively small with less than 1 million parameters (i.e., approximately 100 times smaller than Prithvi). Presto does not convolve over patches to apply attention between spatial patches. Therefore, the authors of Presto state that for image-based predictions, which are ``highly dependent on the occurrence of objects in subregions of the image, models which natively process this important spatial information may be better suited'' \cite[][p. 17]{tseng2023lightweight}. For these reasons, we regard Presto as a complementary approach to Prithvi that is highly effective in leveraging longer time series of satellite imagery.

As a result, so far, scientists commonly use supervised learning models that are purpose-built for the task at hand. We examine several of these studies dealing with fire scar identification, flood mapping, and crop identification to establish a baseline for comparison with Prithvi and highlight the challenges facing the Earth Science community.

Mapping fire scars is of interest to the Earth Science community for tracking trends in wildfire severity,  agricultural burns, and monitoring post-fire recovery. While machine learning techniques such as random forests \cite{10.1016/j.mex.2022.101741, 10.1016/j.rsase.2020.100324}, texture and spectral analysis \cite{10.1080/19475705.2020.1836037}, and support vector machines \cite{10.4236/jgis.2020.123014} are common in the Earth sciences due to ease of implementation, deep learning methods have been utilized as well. U-Nets trained on either optical imagery alone \cite{bs_unet1} or a combination of optical and radar data \cite{bs_unet2} perform well, achieving maximum kappa and F1 scores of 0.9 and 0.84, respectively. Additionally, ongoing learning mechanisms that alleviate catastrophic forgetting have been implemented to finetune the model with new information as it becomes available, foreshadowing the development of pre-trained multitask models \cite{bs_unet2}. Deep neural networks have also been implemented for burn scar detection, reaching an overall accuracy of 97\% with omission errors reaching 30\% for burned areas \cite{bs_dnn}.

Deep learning techniques for flood detection have been in use since at least 2006 \cite{EarlyFloodDL0} when an Artificial Neural Network was used to detect stream flooding in Hawaii based on gauge observations. In 2015-2016, deep learning was combined with satellite remote sensing for flood detection \cite{EarlyFloodDL1, EarlyFloodDL2}. The introduction of a stacked sparse autoencoder to mitigate the lack of labeled data \cite{EarlyFloodDL1} achieved overall accuracy ranging from 85-96\%. More recently, U-Net architectures have been constructed to detect flooding using synthetic aperture radar (SAR) \cite{Flood_SAR}, optical imagery \cite{Flood_PsoUnet}, or a combination of both \cite{Flood_OpticalSAR}. Combining SAR and optical imagery produces a strong overall accuracy of approximately 95\% and a maximum kappa score of 0.92 \cite{Flood_OpticalSAR}. Using optical imagery from Sentinel-2 and a modified U-Net \cite{Flood_PsoUnet}, the model had a maximum overall accuracy of 93\%, IoU of 0.96, and F1 score of 0.80.

Finally, we examine crop segmentation. Crop segmentation is important to stakeholders for practices such as precision agriculture and yield estimation. Historically, several deep-learning techniques have been applied to this problem including deep semantic segmentation (DSS) networks \cite{CropTypeDSS}, U-Net \cite{CropTypeUNet}, and convolutional neural networks (CNN) \cite{CropTypeCNN}. The DSS trained on worldview RGB imagery successfully delineated cropped areas with kappa scores between 0.90-0.97 \cite{CropTypeDSS}. In the same study, U-Net performed at 0.67-0.92 kappa. Use of a CNN segments crop type with mean IoU scores across all classes ranging from 0.74-0.87 and F1 scores ranging from 0.85-0.97\cite{CropTypeCNN}. Additionally, the use of a semantic segmentation model trained on multi-temporal Sentinel-2 imagery produced an overall accuracy of 85\% \cite{CropTypeSS}. Notably, each of the studies referenced herein relies on labeled data, limiting the scale of the model to what labels can be generated. 

These methodologies require ground inspection and local knowledge, are region-specific, and are challenging to scale without incurring significant costs.
Recognizing that remote sensing in Earth Science presently relies on bespoke models for individual tasks and can be constrained by the availability of ground truth data, we aim to demonstrate Prithvi's ability to adapt to each of these tasks and achieve comparable or superior results while requiring fewer labeled data points.

\section{Data for Pretraining}\label{sec2}
In this section, we describe the HLS-2 data source, our novel approach for stratified sampling of geo-tiles, the preprocessing procedure, and data loading considerations that are relevant for large-scale pretraining of Prithvi.

\subsection{Harmonized Landsat Sentinel-2 Dataset}\label{sec:data-pre-processing}
The Harmonized Landsat Sentinel-2 (HLS) project stems from research at NASA's Goddard Space Flight Center in Greenbelt, MD \cite{HLSpub}. The project uses data from NASA/USGS's Landsat 8 and 9 and the ESA's Sentinel-2A and Sentinel-2B satellites to produce a harmonized surface reflectance data product. This data offers observations approximately every two to three days. The creation of the HLS data is an outcome of the Satellite Needs Working Group's assessment in 2016. During this assessment, federal agencies and users identified a demand for frequent Landsat-like observations. These observations help in monitoring short-term vegetation changes and other land elements. The goal is to support agricultural monitoring and detailed land cover classification, encompassing both visible and thermal segments of the electromagnetic spectrum.

Given the spectral similarities among the Landsat 8 Operational Land Imager (OLI), the Landsat 9 OLI-2, and the Sentinel-2 MultiSpectral Instrument (MSI), it is possible to harmonize their data, and this harmonization provides more regular imagery products vital for monitoring the land surface. HLS consists of two data products: The L30 product, derived from Landsat 8 and 9 data, and the S30 product, derived from Sentinel-2 data. Both products can be accessed via Earthdata Search, and NASA's Land Processes Distributed Active Archive Center \cite{HLSL30, HLSS30}. 
HLS data products enhance the current public remote sensing capabilities for land monitoring, especially considering observation frequency. The integration of HLS ensures that the data from both Landsat 8 and 9 (at 30-meter resolution and 16-day repetition) and Sentinel-2A/B (at 10 to 20-meter resolution and five-day repetition) can function as a unified collection. With HLS, observations of the land surface can now be captured at a remarkable 30-meter resolution approximately every two to three days.

The HLS L30 component consists of the surface reflectance derived from all Landsat-8 L1T products and includes nadir adjustment.  The data was accessed and downloaded as cloud-optimized GeoTiffs using the ''HLS Subsetting, Processing, and Exporting Reformatted script''~(HLS-SuPER)~\cite{hls_super} tool from NASA's Land Processes Distributed Active Archive Center~(LP DAAC). Each HLS tile is identifiable by its UTM zone and latitude band. Tiles have dimensions of 3,660$\times$3,660 pixels, which is 109.8$\times$109.8 km$^2$ in area, and slightly overlap each other. While HLS provides up to 15 imaging channels ranging from visible to thermal infrared bands, only 6 channels are used in training the model. These are Sentinel bands 2, 3, 4, 8A, 11, and 12. While the Sentinel band naming convention is used throughout the manuscript, these channels are available in both L30 and S30 products.

The HLS dataset is selected for this FM due to its large archive (3.61 Petabytes) reaching back to 2015 and its breadth of users. HLS does not only serve government interests; in fact, a striking 90\% of its user base is non-government entities. Among these, the education sector stands as a predominant benefactor, constituting 65\% of the user base and utilizing 82\% of the data, which underscores the educational and research merits of HLS. Furthermore, commercial and non-profit users are 10\% and 8\% of the user base, respectively, demonstrating its broader societal and economic impact. Its global reach is another facet, with extensive utilization across diverse geographic locations, including the USA, China, India, Brazil, Columbia, Canada, Germany, Chile, and the UK as top 10 users, marking it as a vital resource for international research \cite{LPDAAC}. The HLS dataset's available bands, multi-sector user base, and global engagement not only amplify its importance but make it an indispensable asset for a myriad of applications for many different users, laying a strong rationale for its consideration in foundational research and beyond. HLS data is being actively used for many different downstream applications, i.e. National-Scale Grassland Management Using HLS  Data in Germany \cite{griffiths2020towards}, Ephemeral Floods Detection in Southeast Australia using HLS Data \cite{tulbure2022can}, Biomass Estimation for Semiarid Rangeland Management in the US \cite{kearney2022monitoring}, Wetland Dynamic, Grassland Fire, and Phenology Monitoring in Central US \cite{zhou2019monitoring}, Disaster response and recovery efforts \cite{physnews}, and more.

\subsection{Efficient Data Sampling}\label{sec:CONTUSsampling}

Due to redundancy in satellite data across regions and years, it is inefficient to pre-train foundation models on all available data. Instead, pretraining requires a representative sample that avoids biases toward the most common landscapes. For example, random sampling may result in biases by focusing on prevalent landscapes while ignoring less common ones. To overcome this challenge, we have implemented a flexible stratified sampling procedure based on aggregate geospatial statistics to allow the sampling of a diversified data set for pre-training. Our algorithm makes use of an efficient querying procedure using overview layers of geospatial-temporal data sets~\cite{freitag2022efficient}. 

\begin{figure}[htb]
    \centering
    \includegraphics[width=0.8\textwidth]{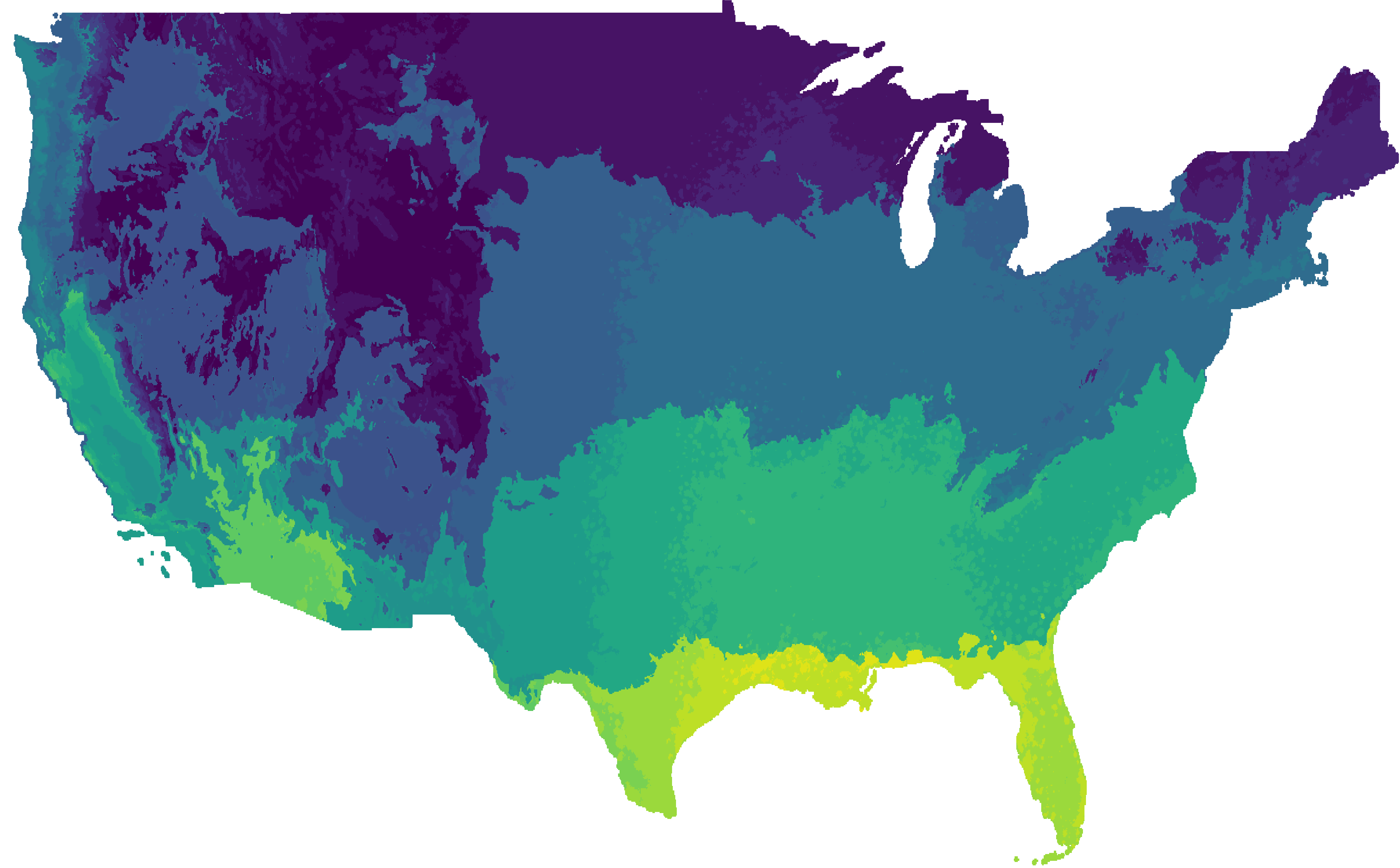}
    \caption{Geo-regions from the contiguous U.S. are clustered into one of 20 different categories based on temperature and precipitation data.}
    \label{fig:us_20blocks}
\end{figure}

Our method follows the following steps: first, we automatically aggregate various geospatial statistics at a low resolution over the whole targeted area. Second, our method divides the low-resolution tiles into several groups based on these statistics. Third, we perform uniform sampling from these groups to ensure a suitable representativeness.

Figure~\ref{fig:us_20blocks} shows an example of groups in the contiguous United States based on temperature and precipitation statistics. To generate this example, firstly, we used our method to compute several statistics on daily mean temperature and precipitation data from 1980 to 2022 at 4km resolution. Both variables were obtained from the PRISM dataset~\cite{PRISM}. Secondly, we divided the contiguous United States into 20 regions considering the mean temperature and precipitation 99\% percentile values aggregated over the 42-year period. Lastly, samples from these regions were sampled to generate a dataset for pre-training, guaranteeing diversity with respect to temperature and precipitation statistics, which can serve as a proxy for different types of landscapes.

\subsection{Preprocessing Routines}
\label{sec:preprocdata}

In contrast to natural images or commonly employed remote sensing benchmark datasets, the large-scale preprocessing of raw satellite images requires a range of additional workflows. We highlight the most relevant ones in the following.

HLS data can contain significant portions of clouds or no-data values. To ensure high-quality data for pre-training, we developed preprocessing routines that exclude images with missing values or containing cloud coverage.
For this, we leverage the cloud mask file (so-called \textit{FMask}), which is associated with each HLS tile at each available date. It contains information regarding cloud coverage categories (cloud, cloud shadows, adjacent to clouds/cloud shadows, no-data) per pixel~\cite{claverie2017harmonized}. We can then use the cloud mask to determine which regions from the tiles have valid data and are cloud-free. 

The evaluation of the images with the help of cloud masks is straightforward. However, this must be conducted for all tiles and timestamps in the sampled dataset. Given our large-scale datasets, this procedure can become significantly time-consuming, which implies that it is not viable to run during training. Therefore, we developed a method to evaluate and filter the dataset \textit{offline} (e.g., before training), keeping track of indices of good quality after preprocessing. In detail, considering a training image size of X$\times$Y pixels, we first divide each sampled M$\times$M HLS tile into non-overlapping windows of size X$\times$Y. Then, we compute the percentage of pixels containing missing values and clouds in each X$\times$Y sub-region of the tile. We define a threshold to filter out sub-regions that do not respect the conditions, keeping only the indices of selected sub-images. The final file contains a list of indices with all the necessary information to identify the regions: tile name, timestamp, and (x, y) coordinates from the top-left pixel of the sub-region. 

To address the efficiency of data preprocessing, we implemented a cloud-native pipeline that can run several processes in parallel. Upon data arrival in our cloud object storage, we compute and store the indices file and the data statistics required for pre-training (mean and standard deviation). The final preprocessing step involves saving the training data in \textit{zarr} files, as explained in the following.

\subsubsection{Zarr preprocessing} \label{sec:zarr}

Once we have the good-quality indices computed, we can directly use this information to load the sub-regions from the GeoTiff files and feed the samples to the model. However, building a single batch of examples for training requires opening multiple files: for a batch size of 128 samples composed of three timesteps and six bands each, we have to open 2304 files. If we need to stream the data for training, this also means that we need to download all those GeoTiff files, even if we only load a portion of them. The high number of file handlers can significantly limit the data loading efficiency, as evidenced in some experiments we conducted. 

Considering these issues, we decided to save the good-quality sub-regions in Zarr files, built only for pre-training purposes. The procedure basically involves loading only the sub-regions listed in the indices file and storing them in Zarr files. The Zarr format supports N-dimensional data and allows read and write arrays concurrently from multiple threads or processes, which makes it appropriate for our pre-training application. Additionally, we store coordinates, timestamps, and other information to identify the sub-regions in the Zarr files. Again, given our large-scale datasets, creating the Zarr files can also be time-consuming. Therefore, we also built a cloud-native method (that can run in multiple processes) to generate the training data.

\section{Model Architecture and Pretraining}
\label{model_arch}

The pretraining of our foundation model is based on the masked autoencoder (MAE) \cite{mae} approach, a successful self-supervised learning method widely used and extended for different data types, including video \cite{videomae} and multi-spectral images \cite{satmae}. The MAE reconstructs masked images using an asymmetric encoder-decoder architecture with a ViT backbone \cite{mae}. 
In this paper, we mainly focus on the 100M version of Prithvi, which has been open-sourced on HuggingFace. 
Each input image is divided into non-overlapping patches of the same size, and a subset of the patches is randomly masked. The encoder receives only the unmasked patches generating their latent representation. The decoder then receives the latent and masked tokens in order to perform the image reconstruction task \cite{mae}.
The pre-training task is the reconstruction of masked tokens, for which the loss function is the mean squared error (MSE) between the masked and predicted tokens in the pixel space, as in the original MAE \cite{mae}.

\begin{figure}[ht]
    \centering
    \includegraphics[width=\textwidth]{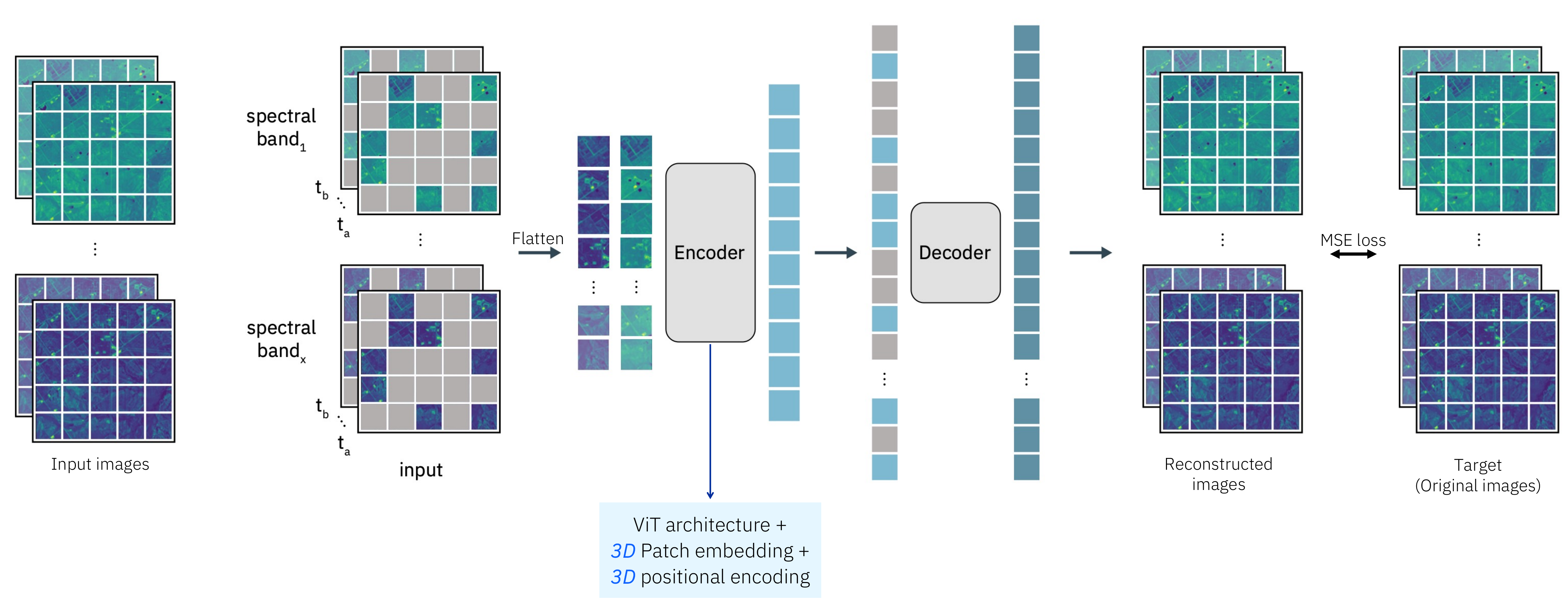}
    \caption{The masked autoencoder (MAE) structure for pre-training Prithvi on large-scale multi-temporal and multi-spectral satellite images.}
    \label{fig:mae}
\end{figure}

In Figure~\ref{fig:mae}, we show the overall MAE structure for pre-training in Prithvi. As shown in the figure, we modify the original MAE to support inputs with temporal and multi-spectral characteristics. Unlike videos, satellite data can be acquired at non-regular and relatively low frequency in time (days) and have more channels than the three used in RGB images. Our main modifications to the ViT architecture are the 3D positional embedding and the 3D patch embedding, which are required to deal with the spatiotemporal data. We describe these modifications in the following.

\subsection{Spatiotemporal Considerations}

Training geospatial AI models require adequately processing 3D spatiotemporal data. We account for this by adjusting the positional embeddings and the patch embeddings in the ViT architecture. 
There are several approaches in the literature for positional encoding considering 3D data (i.e., two-dimensional space plus time). For example, in \cite{spatiotemporalmae}, the authors use separate encodings for temporal and spatial dimensions, while \cite{satmae} creates a relative temporal encoding based on the timestamps of the data. In contrast to previous works, we propose a simple yet effective approach by expanding the 2D sine-cosine positional encoding from the original MAE to a 3D version. For this, we first generate the 1D version of the sine-cosine positional encodings individually for height, width, and time and then combine the individual encodings into a single, 3D positional encoding. We refer to our implementation for details\footnote{https://github.com/NASA-IMPACT/hls-foundation-os}.
For the patch embeddings, we follow work on MAEs for video processing (\cite{spatiotemporalmae} and \cite{videomae}) and leverage 3D patch embeddings. Instead of creating 2D patches for each time step and stacking them in a sequence of tokens (see \cite{satmae}), we divide the 3D input into non-overlapping, equal-sized cubes of data. We specifically make use of 3D convolutions to process the data. In our implementation, we set the tubelet size (i.e., the size of the tube on the temporal dimension) to 1 due to the relatively low frequency of the data.

\subsection{Pretraining}

We use AdamW optimizer with $\beta_1 =$ 0.9, $\beta_2 =$ 0.999, batch size of 1024, one-cycle cosine learning rate scheduler, with a maximum learning rate of 5e-4. We experimented with ViT-base and ViT-large backbones and trained the models for 1000 epochs. The input image size is 224 $\times$ 224, and the patch size is 1 $\times$ 16 $\times$ 16 (time $\times$ x $\times$ y). We used a total of six HLSL30 bands in pre-training, namely B02, B03, B04, B05, B06, and B07.
All the pre-training runs were conducted in the IBM watsonx platform using up to 64 NVIDIA A100 GPUs. 
It is important to highlight that when training models with large-scale spatiotemporal data, the data loading can become a significant bottleneck. Each sample is considerably larger than word tokens, for example, and the overall data size is substantial. Therefore, naive data loading approaches can affect training efficiency and limit GPU usage. 
As mentioned in Section~\ref{sec:zarr}, we save our training data in Zarr files. The data loading scheme leverages the native PyTorch DataLoader, along with \textit{xarray} and \textit{dask}, to load the spatiotemporal samples during training. Our experiments in Table~\ref{tab:epoch_time} show that by adopting this method, we improve our data loading efficiency partially based on a reduction in data required to move to the training environment, as only the filtered samples are stored. Table~\ref{tab:epoch_time} shows the epoch time we registered for the Zarr loading scheme. Compared to the GeoTiff loader, in which we need to open several files and build the samples, we can see that the Zarr loader is significantly faster. Using one-fourth of the number of workers, our Zarr-based approach reduces the epoch time by over 40\% on 8 GPUs and is comparable to GeoTiff loading on 64 GPUs. Based on our data loading approach, our framework fosters large-scale pretraining of geospatial foundation models.

\begin{table}[ht]
    \centering
    \begin{tabular}{lcccc}
    \toprule
         & batch/GPU & workers & prefetch & epoch avg time (s) \\\midrule
        GeoTiff 64 GPUs & 16 & 1 & 2 & 384 \\
        GeoTiff  8 GPUs & 128 & 8 & 2 & 690 \\
        \textbf{Zarr 8 GPUs} & 128 & 2 & 4 & \textbf{381} \\
        \bottomrule
    \end{tabular}
    \caption{Average epoch time in seconds for different runs of data preprocessing and loading. Zarr-based data loading is approximately two times faster than corresponding GeoTiff loading.}
    \label{tab:epoch_time}
\end{table}

\section{Downstream Tasks}
\label{downstream}
We leverage the pretrained foundation model across several downstream tasks. In this paper, we focus on multi-temporal cloud gap imputation, the segmentation of the extent of floods, the segmentation of wildfire scars from wildfires, and multi-temporal crop segmentation. In the following, we introduce the tasks and the underlying data and then focus on modeling to apply the pretrained MAE model to downstream tasks.

\subsection{Datasets}
\noindent
\textbf{Multi-Temporal Cloud Gap Imputation.\footnote{\url{https://github.com/ClarkCGA/gfm-gap-filling-td/tree/main}}} Imagery from the Harmonized Landsat-Sentinel 2 (HLS) dataset was collected across the Contiguous United States from 2022. As with the data for crop segmentation, each TIFF file encapsulates a 224 x 224 pixel region, with a spatial resolution of 30 meters and six spectral bands from three temporal snapshots stacked together for 18 total channels. The time difference between scenes varies between 1 and 200 days. The initial dataset of 10,000 chips was checked to ensure that each time scene had less than 5 percent cloud coverage and zero missing values. This process yielded a final count of 7,852 image chips evenly distributed across the CONUS.  

\noindent
\textbf{Flood Mapping.} Sen1Floods11 \cite{bonafilia2020sen1floods11} is a surface water data set including classified permanent water, flood water, and raw Sentinel-1 and Sentinel-2 imagery. The dataset consists of 4,831 512x512 chips covering 120,406 km\textsuperscript 2 and spans all 14 biomes, 357 ecoregions, and 6 continents of the world across 11 flood events. The benchmark associated with Sen1Floods11 provides results for fully convolutional neural networks trained in various input/labeled data setups, considering Sentinel 1 and 2 imagery. They also provide results achieved by simply thresholding radar backscatter to identify surface water. The dataset and a reference training and evaluation code are available on GitHub\footnote{\url{https://github.com/cloudtostreet/Sen1Floods11}}.

\noindent
\textbf{Wildfire Scar Mapping.\footnote{\url{https://huggingface.co/datasets/ibm-nasa-geospatial/hls_burn_scars}}} We gather data on wildfire scars using the Monitoring Trends in Burn Severity (MTBS) historical fire database. The database is created by the United States Geological Survey Center for Earth Resources Observation and Science (EROS) and the United States Department of Agriculture Forest Service Geospatial Technology and Applications Center (GTAC). The database contains shapefiles for both wildfires and intentional burning from 1984 to the present; however, in this work, the time period is limited to the years 2018--2021 due to the availability of Harmonized Landsat Sentinel-2 (HLS) observations. When selecting HLS imagery for each fire, the first HLS image that occurs more than 1 month but less than three months after the start of the fire is chosen. This is to ensure that the fire has been fully contained and reached its full extent in the satellite imagery, but vegetation has not yet had time to recover. If the first HLS scene has more than 10\% cloud cover or the satellite overpass captures less than 50\% of the tile, that image is discarded and replaced with the earliest that meets these criteria.  Once an HLS image is located, it is subsetted to a 512x512 pixel image centered on the wildfire scar. A corresponding GeoTiff mask is created from the fire shapefile. After visual inspection to ensure visibility of the wildfire scar and cloud-free conditions, 805 scenes are available for training and validation.

\noindent
\textbf{Multi-Temporal Crop Segmentation.\footnote{\url{https://huggingface.co/datasets/ibm-nasa-geospatial/multi-temporal-crop-classification}}} Imagery from Sentinel satellites depicting various land cover and crop types across the Contiguous United States was gathered for the year 2022, with the categorization labels sourced from USDA's Crop Data Layer (CDL). This dataset primarily aims to train geospatial machine learning models for segmentation tasks. 
Each TIFF file encapsulates a 224 x 224 pixel region with a spatial resolution of 30 meters. Every input satellite file includes 18 bands, with six spectral bands from three temporal snapshots stacked together. The corresponding GeoTiff mask file holds a single band designating the target classes for each pixel.

In building this dataset, an initial batch of 5,000 chips was outlined based on samples from the USDA CDL to ensure a balanced representation across the CONUS. Following this, each chip had its corresponding HLS S30 scenes between March and September 2022 looked up, and scenes with minimal cloud cover were chosen. Three scenes were then picked from the low-cloud scenes to capture different stages of the season - one from the early season, one from the mid-season, and one from the late season. These selected scenes were reprojected to align with the CDL's projection grid (EPSG:5070) using bilinear interpolation. Subsequently, the three scenes for each chip were trimmed to the chip's bounding box, and the 18 spectral bands were stacked. Additionally, a quality check was carried out on each chip using the Fmask layer from the HLS dataset, discarding any chip with clouds, cloud shadows, proximity to clouds, or missing values. This process yielded a final count of 3,854 chips.

\begin{figure}[htb]
    \centering
    \includegraphics[width=0.95\textwidth]{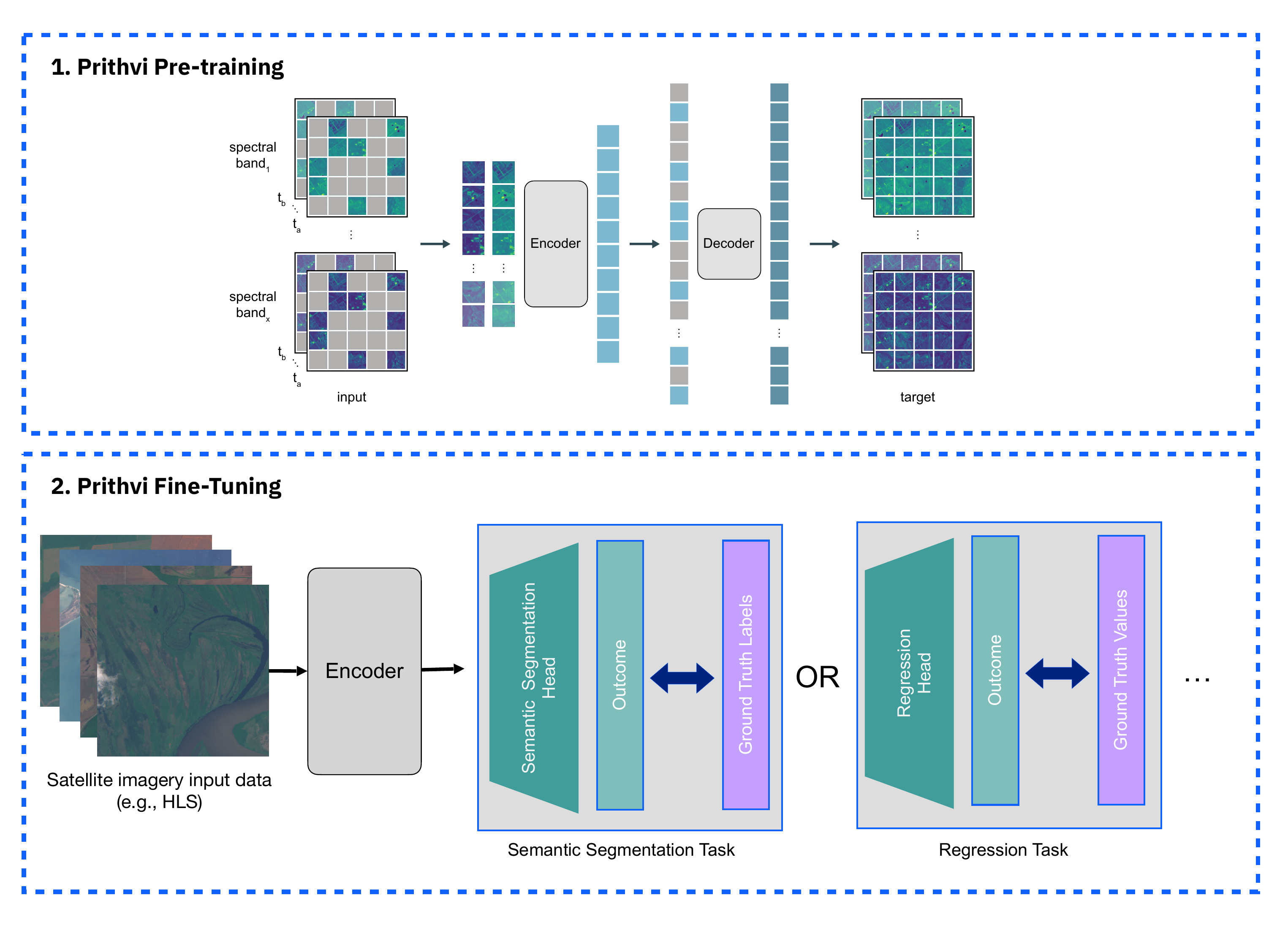}
    \caption{Pre-training and fine-tuning in Prithvi for various types of downstream tasks.}
    \label{fig:tasks}
\end{figure}

Binary HLS cloud masks were taken from the Fmask channel to use as synthetic cloud masks for the purposes of fine-tuning and inference, where cloudy pixels are represented as 1 and non-cloudy pixels as 0. Cloud masks are from identical regions of the United States as the image chips and encompass a 224 x 224 pixel region with a spatial resolution of 30 meters. 21,648 binary cloud masks were generated through this process.

\subsection{Downstream Modeling Considerations}

Our model has been pre-trained in a masked autoencoder scheme that learns to reconstruct the masked input images using the encoder features, as shown in Figure~\ref{fig:tasks}.1. In the fine-tuning module, we reuse or fine-tune the encoder pre-trained weights and learn decoder weights for specific tasks (Figure~\ref{fig:tasks}.2). Our fine-tuning module is based on Pytorch and builds on the mmsegmentation \cite{mmseg2020} library, which natively supports semantic segmentation tasks. 
To meet the requirements of complex spatiotemporal remote sensing data, we have enhanced the functionalities of mmsegmentation to deal especially with spatiotemporal data and have customized the library to other tasks such as regression and classification. In this paper, we will focus on segmentation capabilities of the enhanced library. 

For the task-specific decoder heads, we propose a lightweight architecture to facilitate fine-tuning. Specifically, our approach utilizes a neck composed of four ConvTranspose2D layers and, ultimately, a single, two-dimensional convolutional layer. 
We define the hyperparameters for the fine-tunings based on a range of pilot experiments. In our final experiments, we employ a batch size of 4 for flood mapping and wildfire scars (8 for multi-temporal crop segmentation), fine-tune over a maximum of 60 epochs for flood mapping (50 for wildfire scars, 80 for multi-temporal crop segmentation), and a learning rate of 6e-5 for flood mapping (1.3e-05 for wildfire scars, 1.5e-05 for multi-temporal crop segmentation). We utilize weighted cross entropy loss to account for class imbalance for flood mapping and multi-temporal crop segmentation (unweighted dice loss for wildfire scars. 
We evaluate the segmentation tasks based on the Intersection-over-Union (IoU) metric, F1-score, and Accuracy following previous work \cite{mmseg2020}. The mean values of these metrics are averaged over the number of classes (mIoU, mF1-score, and mAcc). 
For cloud gap imputation, we adapted the pre-training process of Prithvi to perform inference on input cloud masks and fine-tune the encoder from pre-trained weights without a decoder head. In our final experiments, we employ a batch size of 16 and a learning rate of 1e-4 for fine-tuning over a maximum of 200 epochs. During training, we used the Root Mean Squared Error (RMSE) as our loss metric. To evaluate the imputation performance, we employed RMSE, Mean Absolute Error (MAE), and the Structural Similarity Index Measure (SSIM). 

\section{Results}

We comprehensively assess the performance of Prithvi in a wide range of experiments for both its pretraining and its application to four selected downstream tasks. 
For the latter, we focus on multi-temporal cloud gap imputation, as well as segmentation tasks, such as flood mapping, mapping of wildfire scars, and multi-temporal crop segmentation.

\subsection{Pretraining Results}

\begin{figure*}[ht]
    \centering
     \begin{subfigure}[t]{0.35\textwidth}
          \centering
    \includegraphics[width=0.85\textwidth]{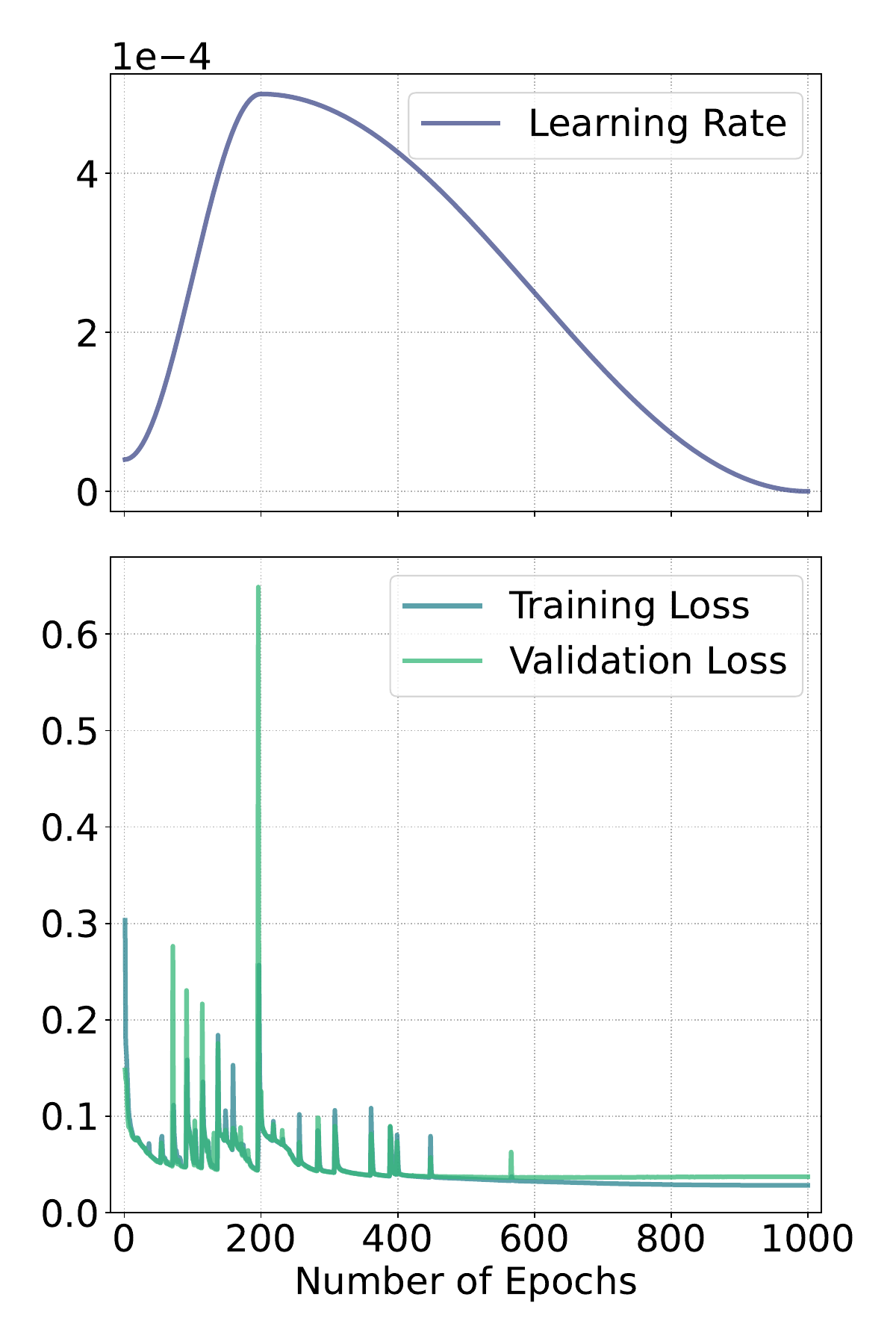}
    \caption{MSE training and validation loss curves during pretraining accompanied by the associated values of the learning rate scheduler. Training loss decreases to 0.0283,  validation loss is lowest at 0.0364.}
    \label{fig:geofm-loss}
     \end{subfigure}
    \centering
    \begin{subfigure}[t]{0.6\textwidth}
         \centering
    \includegraphics[width=\textwidth]{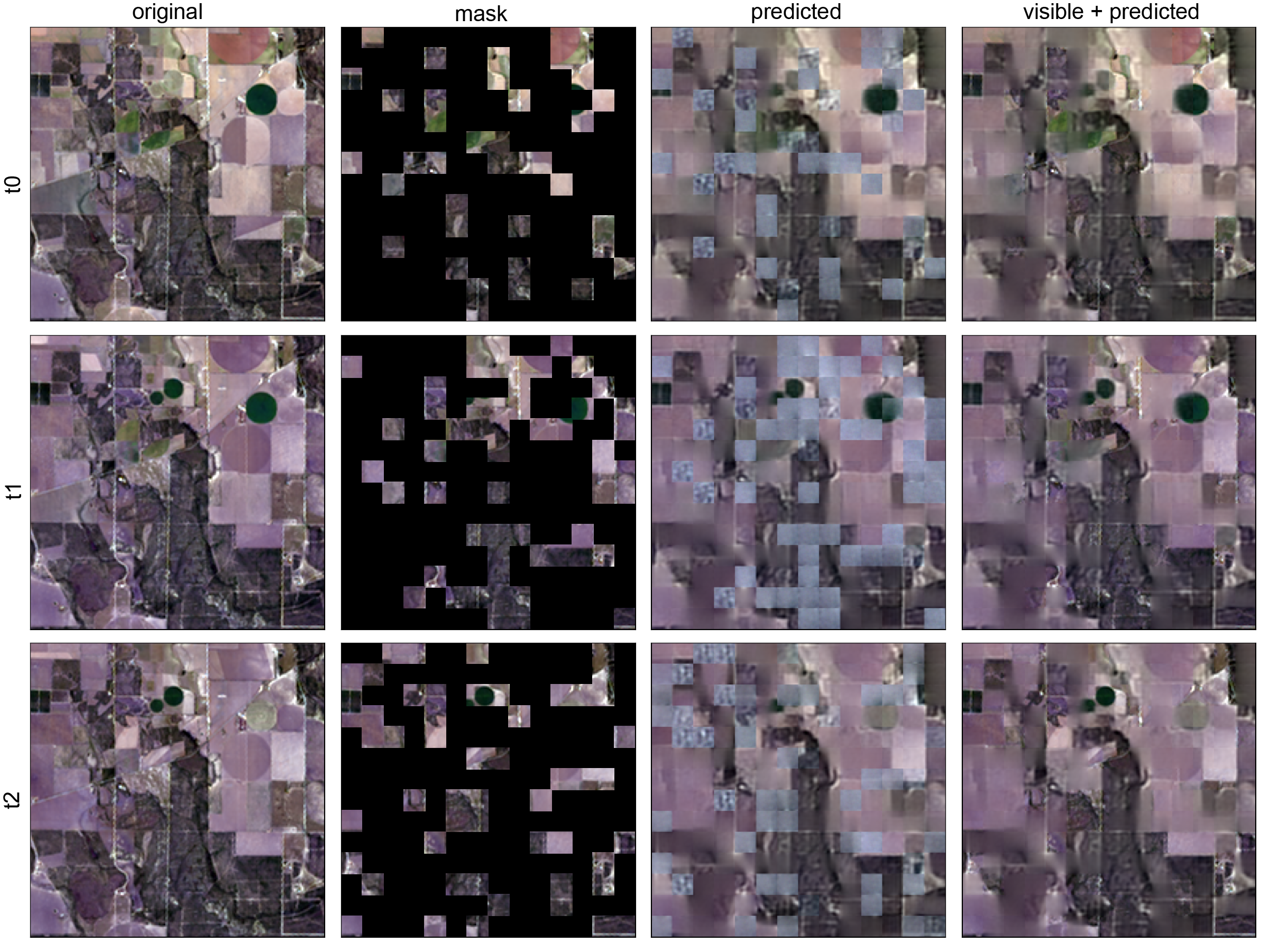}
    \caption{Reconstruction results on images unseen during training (different locations) with Prithvi model with ViT-base backbone. Here we show the RGB bands together (B04, B03, and B02, respectively) for better visualization, although the model also predicts B05, B06, and B07.}
    \label{fig:pretrain_test_rgb}
     \end{subfigure}     
    \begin{subfigure}[t]{\textwidth}
    \includegraphics[width=\textwidth]{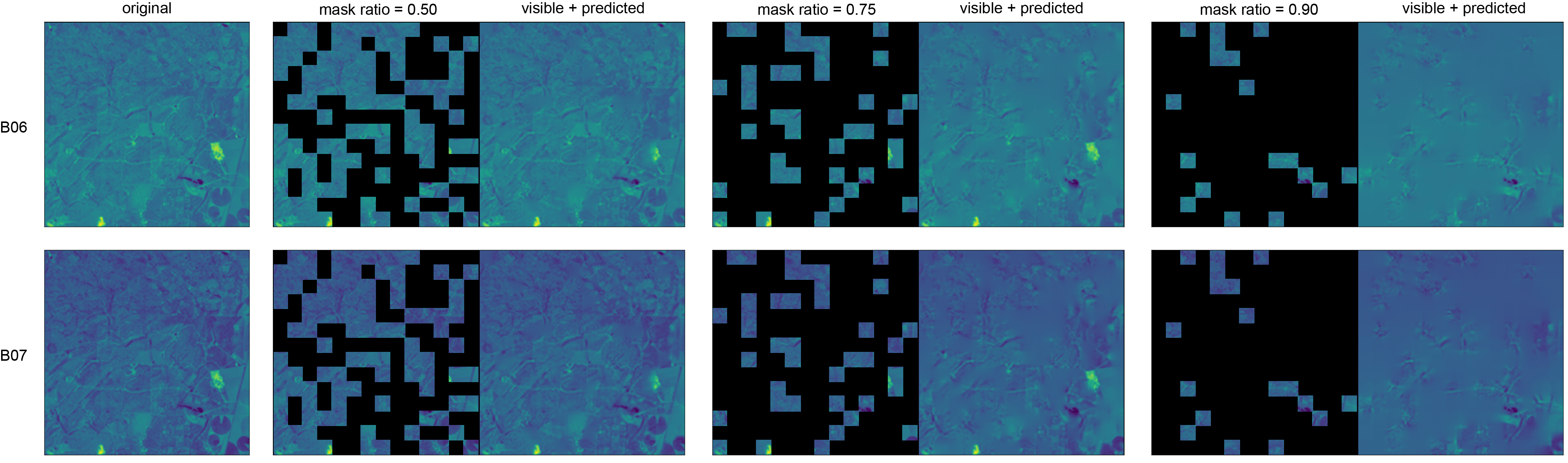}
    \caption{Reconstruction results on images unseen during training (different locations) with Prithvi model for bands B06 and B07 for different masking ratios with ViT-base backbone. Here, we show a single time step of an input image unseen during training.}
    \label{fig:pretrain_test_B6B7}
    \end{subfigure}
     \caption{Pretraining results of Prithvi using 1TB of HLS data from the contiguous US.}
     \label{fig:pretraining-results}
\end{figure*}

We present the loss of the reconstruction during pretraining in Figure~\ref{fig:geofm-loss}. Overall, we observe a consistent convergence, where peaks in the loss can be attributed to updates in the learning rate. The model achieves a minimal training MSE loss of 0.0283, which is accompanied by a validation loss of 0.0364. 
During pretraining, the model learns to reconstruct masked pixels in the input data. To qualitatively validate our pretraining, we depict visualizations of reconstructions on test data (unseen regions) in Figure~\ref{fig:pretrain_test_rgb} and Figure~\ref{fig:pretrain_test_B6B7}.
In Figure~\ref{fig:pretrain_test_rgb}, we focus on the temporal aspects of reconstruction. 
Therefore, we particularly focus on parts of the image that change over the three depicted timesteps---such as the circular shape in the upper right of the images. 
While most of this circular shape is masked across all three timesteps, the model successfully predicts a change in the color of the shape in \textit{t2}.
This visually confirms the temporal abilities of reconstruction.
In Figure~\ref{fig:pretrain_test_B6B7}, different masking ratios are depicted to visualize their effect on the reconstruction result. 
We observe that when half of the input image is randomly masked, the reconstruction of pixels is highly accurate. 
As expected, this ability declines with increasing masking ratio. 
However, even with 90\% of the input image masked, the reconstruction around visible pixels is accurate (e.g., see the lower-right area in the third column of images).
Overall, we observe that the model successfully reconstructs RGB bands and infrared bands across time and for different masking ratios.

\subsection{Downstream Application Results}

In the following, we present results from applying Prithvi to multi-temporal cloud gap imputation, the segmentation of floods and wildfire scars, as well as multi-temporal crop segmentation. 
The code and demos for these use cases have also been open-sourced by NASA and IBM and can be found on HuggingFace\footnote{https://huggingface.co/ibm-nasa-geospatial} and GitHub\footnote{https://github.com/NASA-IMPACT/hls-foundation-os}.

\subsubsection{Multi-Temporal Cloud Gap Imputation}
For the gap imputation task, the encoder of the Prithvi 100M model is modified to accept input masks in the middle time step instead of applying random masks, masking any patches with clouds present. Then, the model is tasked with filling the masked portions of the cloud gap imputation dataset\footnote{\url{https://github.com/ClarkCGA/gfm-gap-filling-td/tree/main}}. 
As a baseline, we used a modified CGAN architecture derived from Pix2Pix \cite{pix2pix2017} building on previous work \cite{baier2020building}. We trained this model from scratch using the same cloud gap imputation dataset. The CGAN model takes known pixel values from three time steps and uses this is a condition to generate masked pixel values in the center time step. A convolutional multi-scale patch discriminator was used, with loss values calculated only for patches with cloudy pixels present. Mean Squared Error (MSE) loss was also calculated for only masked pixels and was used along with the masked discriminator hinge loss to update the generator, with hyperparameter alpha defining the weight given to MSE loss relative to hinge loss. For all experiments using the CGAN model, different learning rates were used for the generator and discriminator \cite{DBLP:journals/corr/HeuselRUNKH17}. The learning rate for the discriminator was 1.0e-4, the learning rate for the generator was 5.0e-4, and alpha was 5, meaning that MSE loss was given 5 times the weight of hinge loss in updating the model. For optimization purposes, Adam optimizer was used with a constant learning rate.

\begin{figure*}[ht]
    \centering
     \begin{subfigure}[t]{0.45\textwidth}
          \centering
    \includegraphics[width=\textwidth]{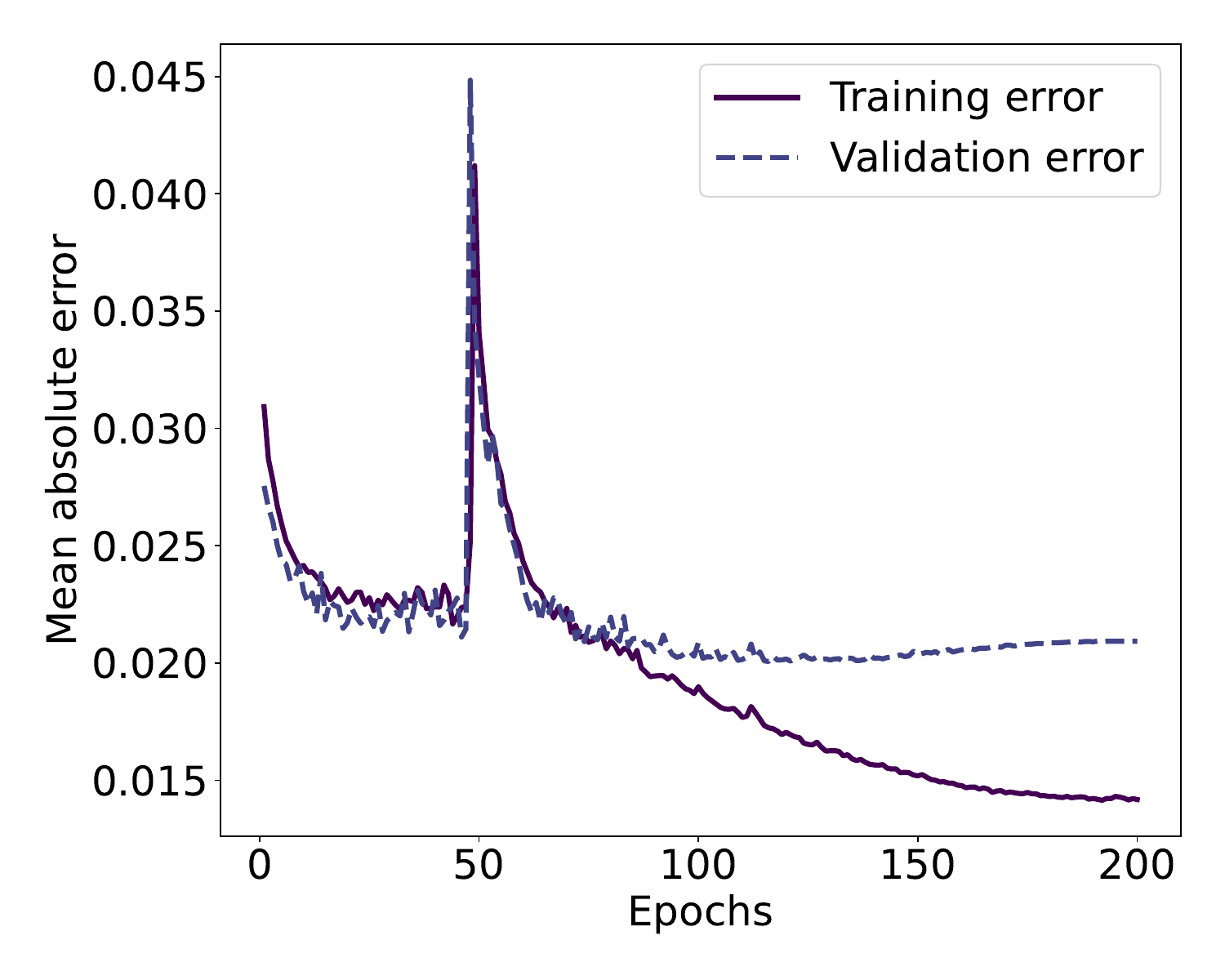}
    \caption{Mean absolute error}
    \label{fig:prithvi_6231_training_graph_mse}
     \end{subfigure}
    \centering
        \begin{subfigure}[t]{0.45\textwidth}
         \centering
    \includegraphics[width=\textwidth]{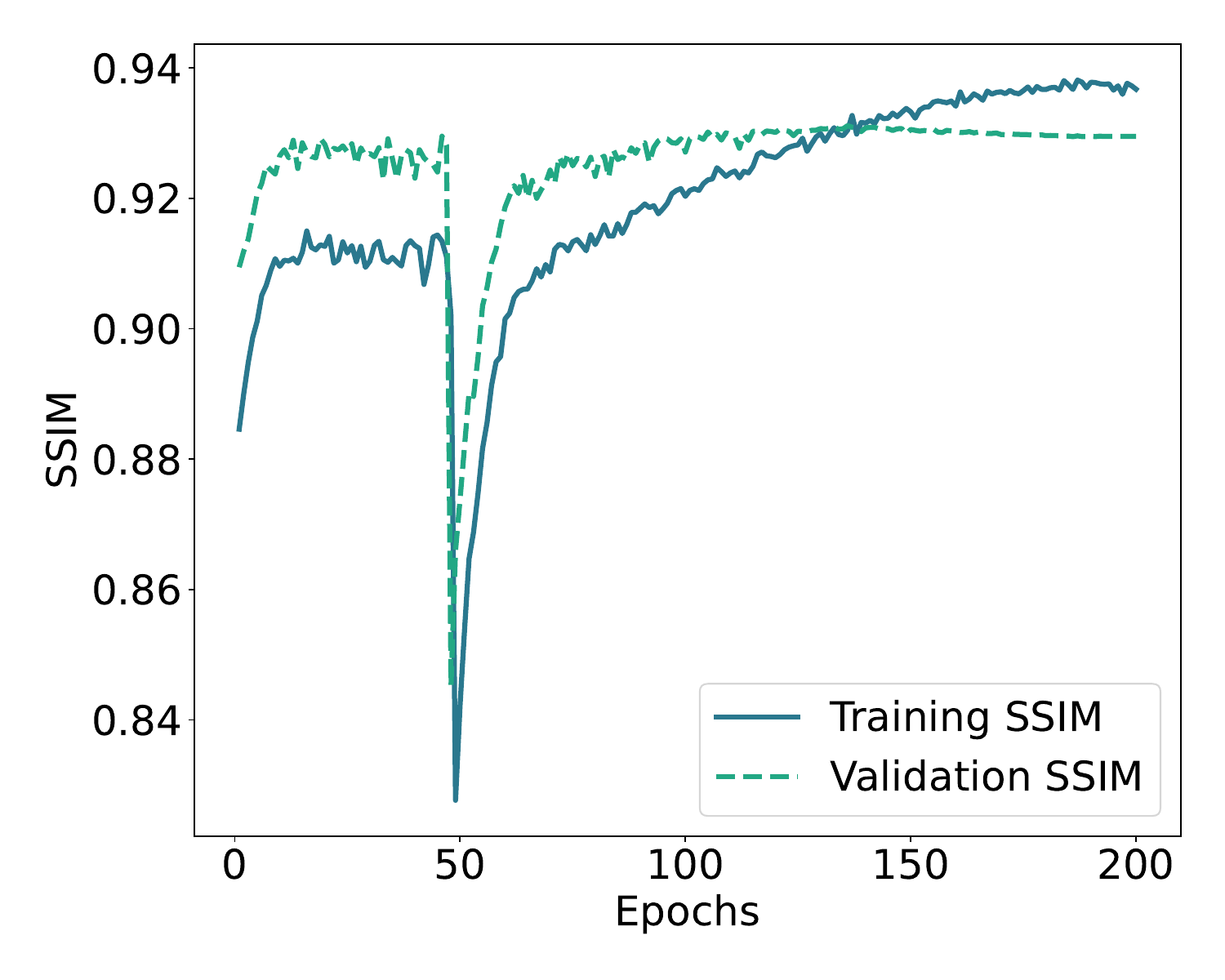}
         \caption{SSIM}
    \label{fig:prithvi_6231_training_graph_ssim}
     \end{subfigure}     
     \caption{Performance of Prithvi during fine tuning for multi-temporal cloud gap imputation.}
     \label{fig:prithvi_6231_training_graph}
\end{figure*}

\begin{figure*}[ht]
    \centering
    \begin{subfigure}[t]{0.45\textwidth}
        \centering
    \includegraphics[width=\textwidth]{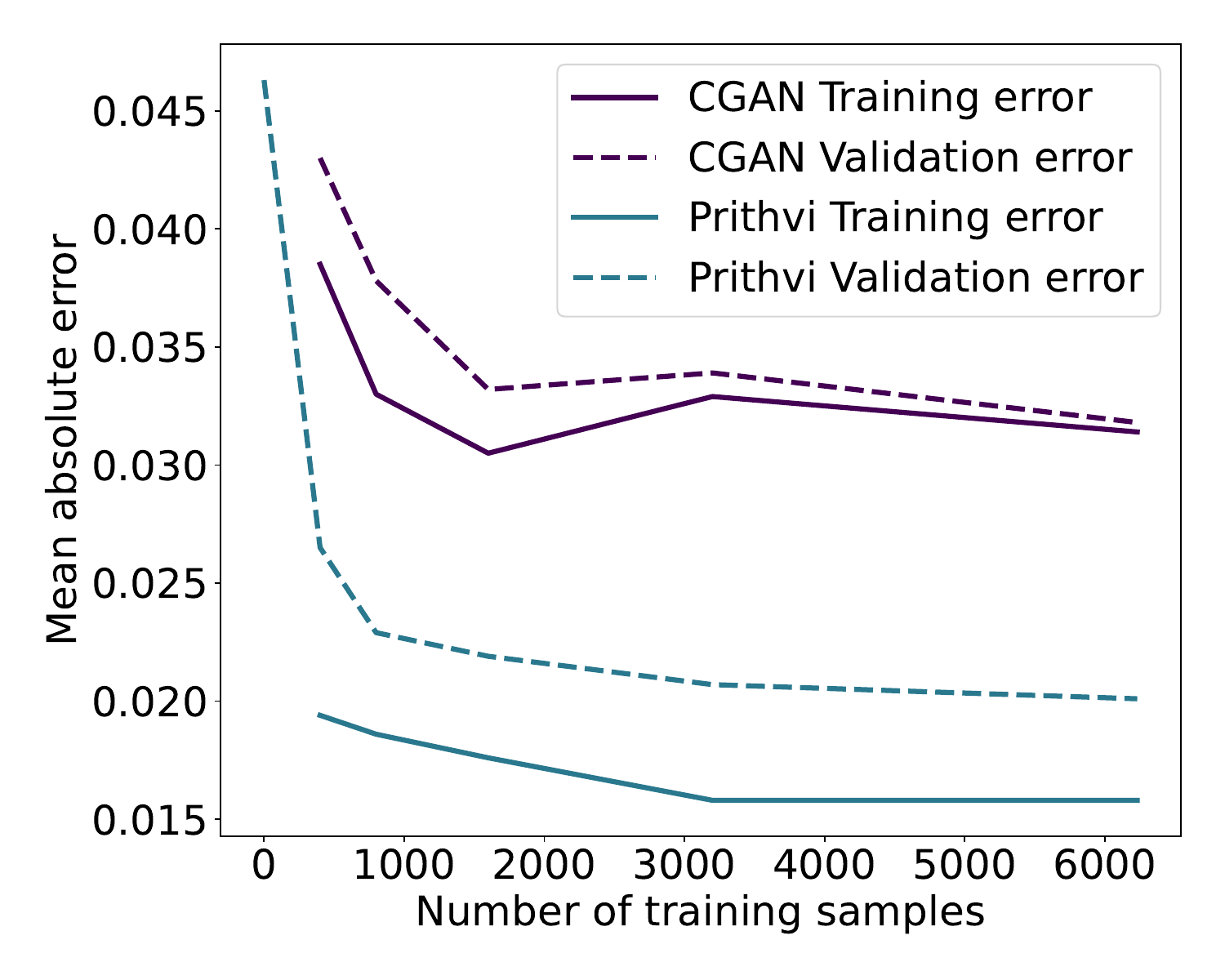}
    \caption{Mean absolute error after 200 epochs.}
    \label{fig:gap_filling_mae}
     \end{subfigure}
    \centering
        \begin{subfigure}[t]{0.45\textwidth}
             \centering
    \includegraphics[width=\textwidth]{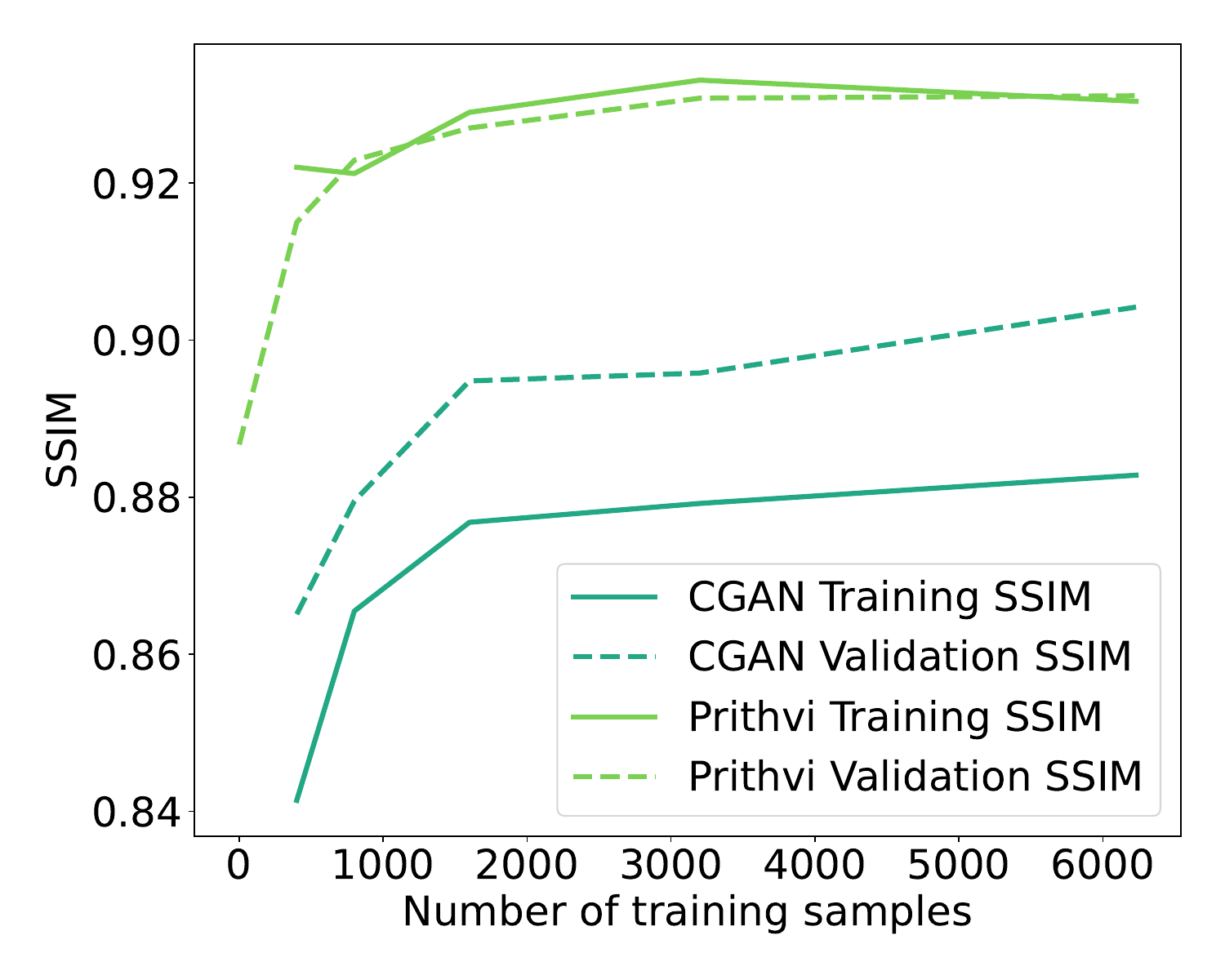}
    \caption{Structural similarity index measure after 200 epochs.}
    \label{fig:gap_filling_ssim}
     \end{subfigure}     
     \caption{Comparison of performance metrics for Prithvi and CGAN on multi-temporal cloud gap filling using a range of training subsamples.
     }
     \label{fig:cloud_loss}
\end{figure*}

We compared model effectiveness by training on subsets of data. First, 1621 image chips were reserved as a consistent validation set across all experiments of both models. 1621 cloud masks were selected for the validation set with a consistent distribution from 1-100 percent coverage. All images and cloud masks were matched identically across all experiments. Then, experiments were run with random subsets of 6231, 3200, 1600, 800, and 400 image chips matched randomly during training to cloud masks.
During fine-tuning of the encoder, Figure~\ref{fig:prithvi_6231_training_graph}~(a) shows that loss converges to an initial minimum within the first 20 epochs, then spikes and converges to a second, marginally lower minimum. Even after a few epochs, Prithvi achieves a SSIM of more than 0.9, which is depicted in Figure~\ref{fig:prithvi_6231_training_graph}~(b). 
In Figures \ref{fig:gap_filling_mae} and \ref{fig:gap_filling_ssim}, we see that Prithvi outperforms the CGAN model in terms of Mean Absolute Error and SSIM consistently across all subsets. Importantly, Prithvi outperforms the CGAN (finetuned on the entire dataset of 6,231 samples) even when only leveraging 400 samples in the finetuning process by 1.2\% in SSIM.
We also visualize results for qualitative comparison. 

\begin{figure*}[ht]
    \centering
     \begin{subfigure}[t]{0.45\textwidth}
          \centering
    \includegraphics[width=\textwidth]{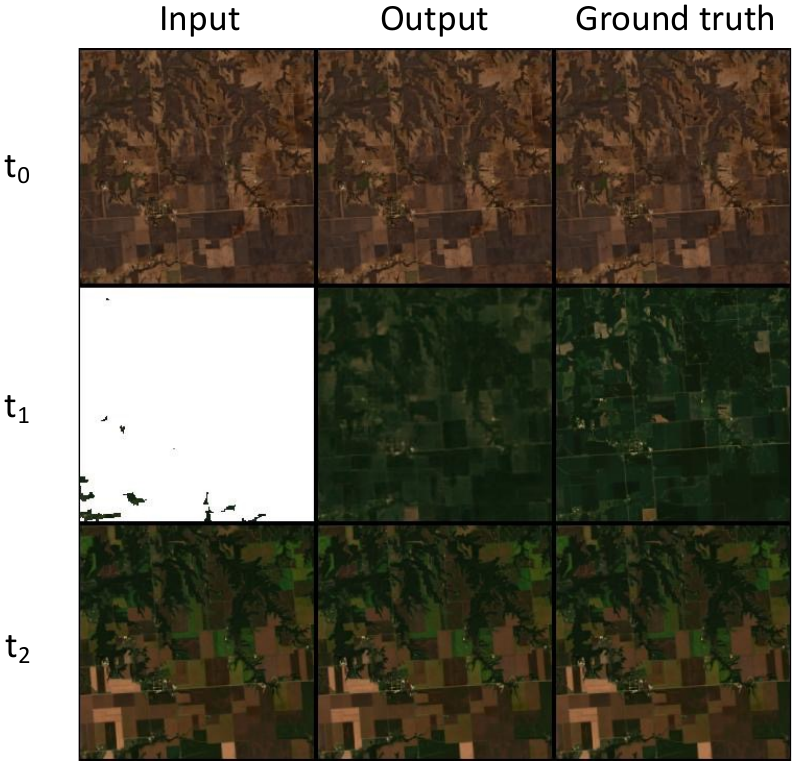}
    \caption{\textbf{Prithvi:} Model can infer pixel values without access to the date of any of the time steps.}
    \label{fig:prithvi_cloud_reconstruction}
     \end{subfigure}
    \centering
        \begin{subfigure}[t]{0.45\textwidth}
         \centering
    \includegraphics[width=\textwidth]{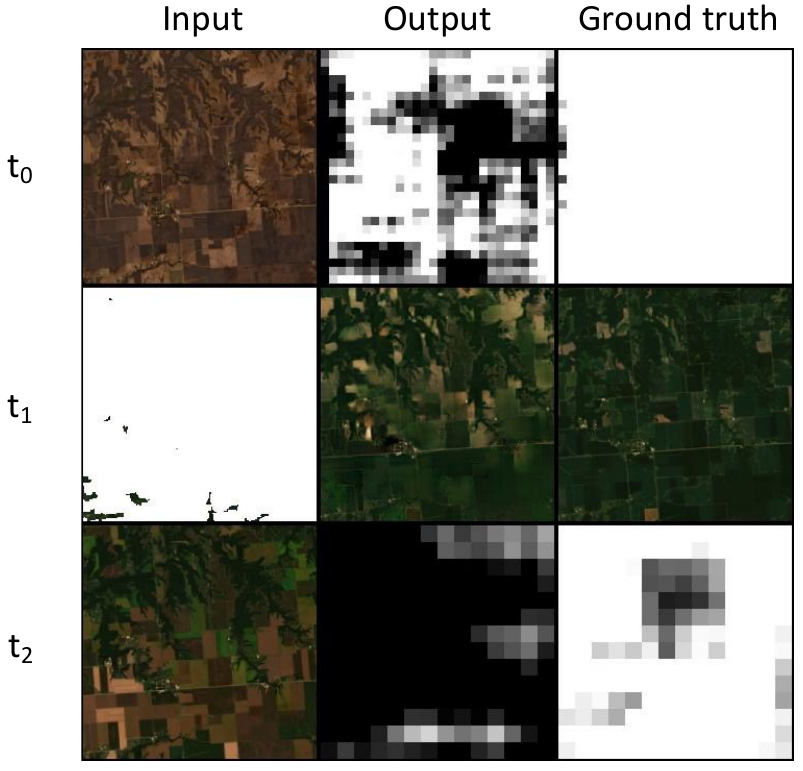}
         \caption{\textbf{CGAN:} The output of the discriminator for the reconstruction and the ground truth is included in place of non-masked time steps.}
    \label{fig:cgan_cloud_reconstruction}
     \end{subfigure}     
     \caption{Comparison of true color composites of model outputs for Prithvi and CGAN for multi-temporal cloud gap filling.  From left to right: model input, reconstruction using model output, ground truth. Each row represents a time step.
     }
     \label{fig:cloud_reconstruction_comparison}
\end{figure*}

In Figure \ref{fig:prithvi_cloud_reconstruction}, Prithvi, after being fine-tuned with the full training dataset, is tasked with imputing missing pixels in the center time step of this image from the validation set, and is able to infer pixel values for an intermediate season without any information on the date of any of the time steps. The output of the CGAN model in Figure \ref{fig:cgan_cloud_reconstruction} given the same inputs shows that it is less effective at constraining pixel values within reasonable bounds. This result is borne out in a comparison of pixel value ranges and spectral band relationships for each model.

\subsubsection{Flood Mapping}

\begin{figure*}[ht]
    \centering
     \begin{subfigure}[t]{0.45\textwidth}
          \centering
    \includegraphics[width=\textwidth]{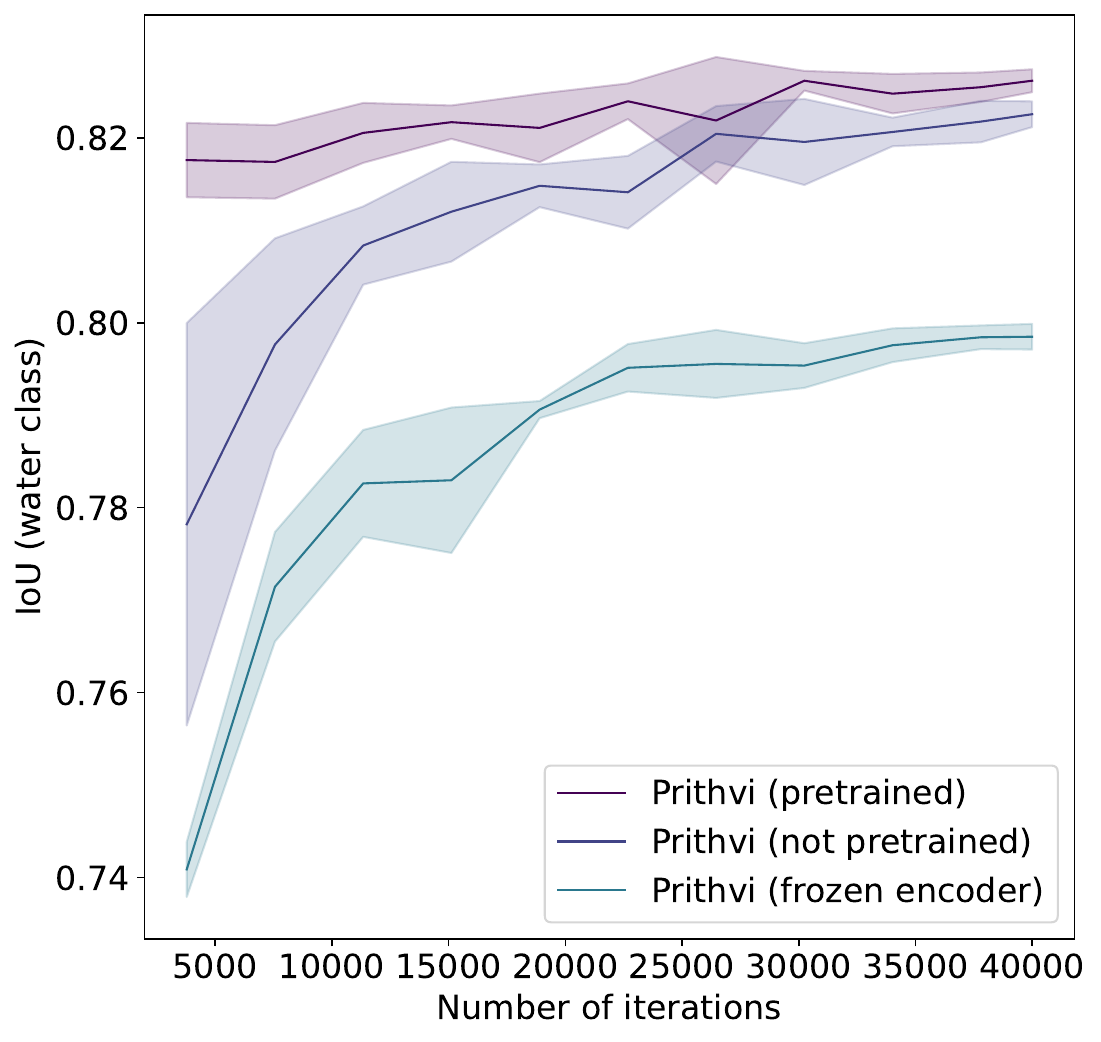}
    \caption{Performance based on (1) pretrained, (2) randomly initialized, and (3) frozen encoder weights. Confidence bands represent the standard deviation across 5 different seeds.}
    \label{fig:lood-mapping-performance}
     \end{subfigure}
    \centering
        \begin{subfigure}[t]{0.45\textwidth}
         \centering
    \includegraphics[width=\textwidth]{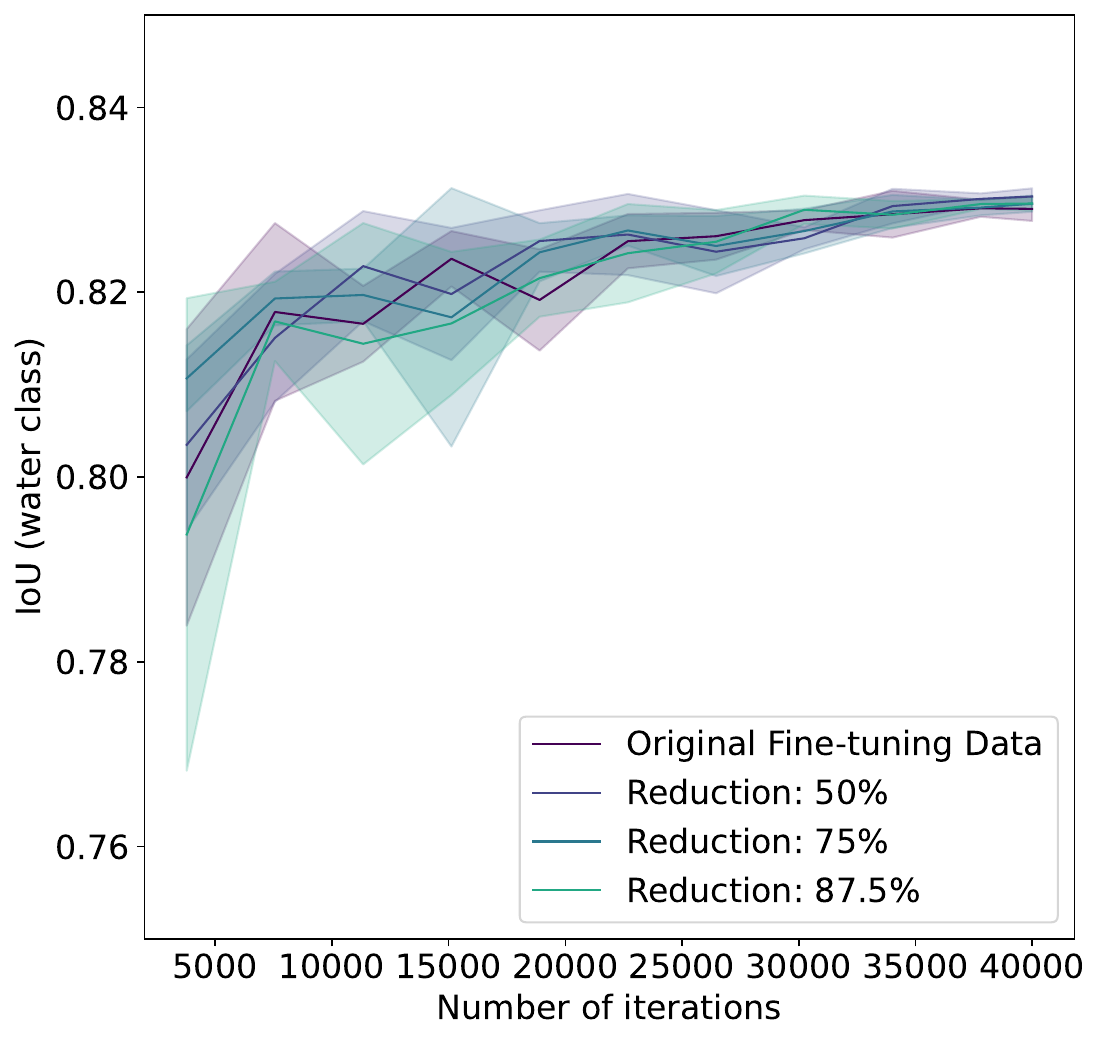}
    \caption{Data efficiency of pretrained Prithvi in terms of reduction of required labeled images for fine-tuning in the flood mapping task using ViT-large backbone.}
    \label{fig:lood-mapping-efficieny}
     \end{subfigure}     
     \caption{Evaluation of Prithvi on Sen1Floods11 test set regarding (a) the performance and (b) the data efficiency using the ViT-large backbone.}
     \label{fig:flood-mapping-results}
\end{figure*}

In the following, we present experimental results on the performance and the data efficiency of Prithvi when generalizing to the different resolutions of S2 data (10m). In Figure~\ref{fig:lood-mapping-performance}, we observe that pretraining Prithvi on HLS data of 30m resolution strongly accelerates the fine-tuning process for flood mapping on 10m resolution compared to the same architecture without pretraining. 
In preliminary experiments, we have found that, on average, the pretrained model surpasses a ViT-base model (see also Table~\ref{tab:flood_mapping_results}) after 25 epochs of fine-tuning, while the same architecture with randomly initialized weights requires 55 epochs. This indicates that pretrained Prithvi accelerates the finetuning process to achieve this reference performance by more than factor two. Overall, the experiments in Figure~\ref{fig:lood-mapping-performance} demonstrate that pretraining on data from the U.S. accelerates the accuracy of the model even when fine-tuned on data from global flood events and, thereby, emphasizes the generalizability of Prithvi. Additionally, our findings provide evidence for the generalizability of Prithvi towards a differing resolution during finetuning (in this case, the resolution increases from 30m in pretraining to 10m in finetuning).

In Figure \ref{fig:lood-mapping-efficieny}, we demonstrate that Prithvi allows us to reduce the number of required labeled images significantly while preserving performance. For example, in our experiments, we reduce the number of labeled images during fine-tuning by half, from 252 images to 126 images, and observed a very similar performance to fine-tuning on the entire set of labeled training images. Across the eleven geographic regions, 126 images refer to approximately 11 labeled images per geographic region. Even if we decrease the number of labeled images further by close to 90\%, the models still converge to an IoU of over 80\% on average. 

In Table \ref{tab:flood_mapping_results}, we compare the performance of Prithvi for flood mapping with the baseline in \cite{bonafilia2020sen1floods11}, as well as recent off-the-shelf vision transformer architectures (based on standard parameters for ViT and Swin as proposed in \cite{mmseg2020}). After 50 epochs, we observe an IoU on the water class of 81.26, surpassing the performance of off-the-shelf vision transformer architectures (i.e., standard implementations of ViT, Swin) both with random and pre-trained weights. 
Top performance of Prithvi was obtained after 500 epochs, where Prithvi achieves an IoU score of 82.99 on the target class (i.e., water class). We observe that after finetuning over a significant number of epochs, the performance of pretrained and non-pretrained Prithvi converges, which is in line with our expectations. 

\begin{table}[ht]
    \centering
    \resizebox{\textwidth}{!}{
    \begin{tabular}{l|cc|ccc}
    \toprule
         & IoU & F1 & mIoU & mF1-score & mAcc  \\
         & (water) & (water) & (both classes) & (both classes) & (both classes)  \\\midrule
         Baseline \cite{bonafilia2020sen1floods11} & 24.21 & --  & -- & -- & -- \\[5px]

         ViT-base \cite{dosovitskiy2020image} & 67.58 & 80.65  & 81.06 & 88.92 & 88.82 \\
         Swin \cite{liu2021swin} & 79.43 & 88.54 & 87.48 & 93.13 & 90.63 \\ 
         Swin$\dagger$ \cite{liu2021swin} & 80.58 & 89.24 & 87.98 & 93.44 & 92.02 \\[5px]
         
         \textsc{After 50 epochs} \\
         Prithvi (not pretrained) & 80.67 & 89.30 & 88.76 & 93.85 & 94.79   \\
         {Prithvi (pretrained)}  & {81.26} & {89.66} & {89.10} & {94.05} &  \textbf{95.07} \\[5px]

         \textsc{After 500 epochs} \\
         Prithvi (not pretrained) & 82.97 & 90.69 & 90.14 & 94.66 &  94.82  \\
         \textbf{Prithvi (pretrained)}  & \textbf{82.99} & \textbf{90.71} & \textbf{90.16} & \textbf{94.68} & {94.60}   \\[5px]
    \bottomrule
    \end{tabular}
    }
    \caption{Prithvi performance compared to baseline of Sen1Floods11 \cite{bonafilia2020sen1floods11}, as well as recent vision transformer architecture baselines used off-the-shelf with standard hyperparameters from \cite{mmseg2020}. Performance is calculated pixel-wise over the test set, accounting for class imbalance. $\dagger$ Swin pretrained on ADE20K.}
    \label{tab:flood_mapping_results}
\end{table}

\subsubsection{Wildfire Scar Mapping}

\begin{figure*}[ht]
    \centering
     \begin{subfigure}[t]{0.45\textwidth}
          \centering
    \includegraphics[width=\textwidth]{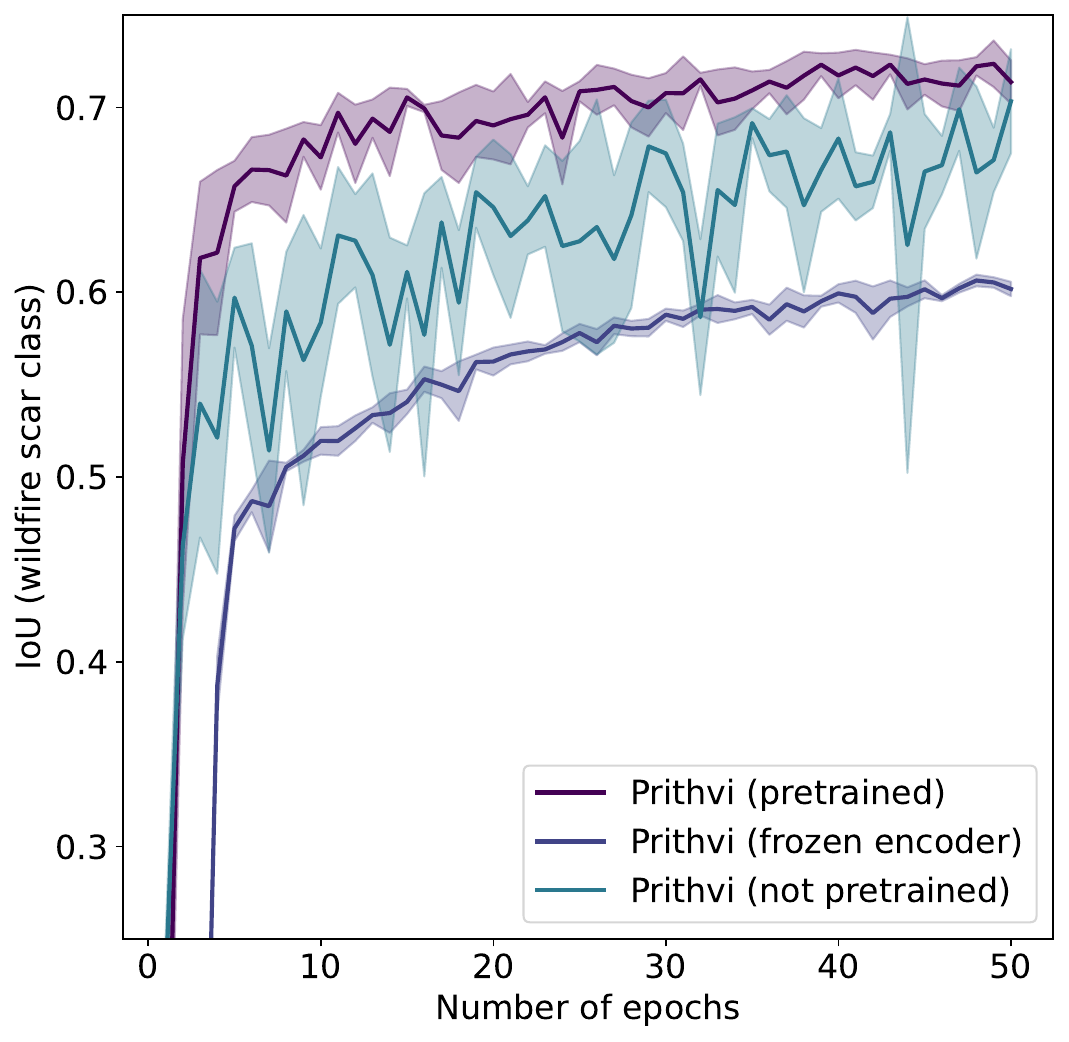}
    \caption{Performance based on (1) pretrained, (2) randomly initialized, and (3) frozen encoder weights. Confidence bands represent the standard deviation across 5 different seeds.}
    \label{fig:fire-performance}
     \end{subfigure}
    \centering
        \begin{subfigure}[t]{0.45\textwidth}
         \centering
    \includegraphics[width=\textwidth]{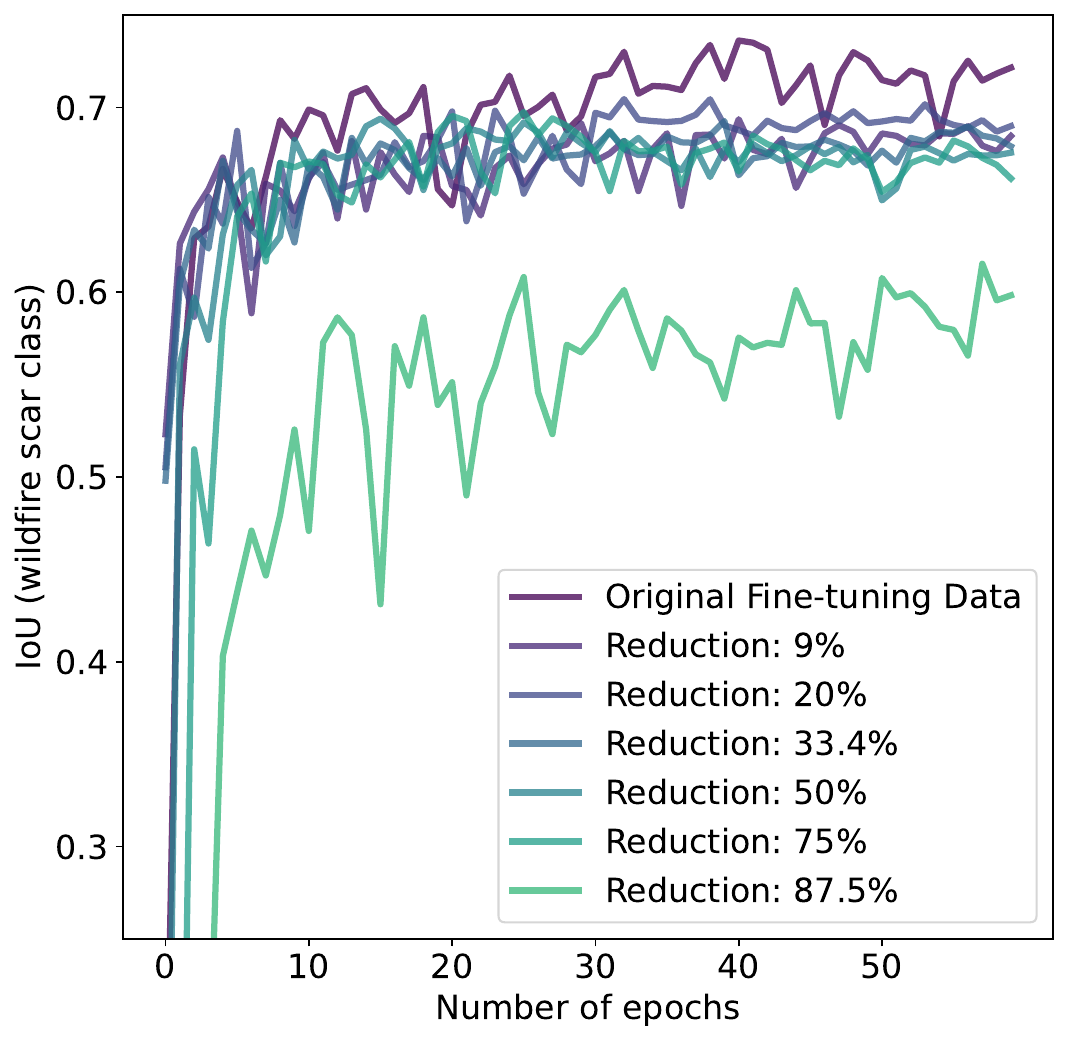}
         \caption{Pre-trained Prithvi: Data efficiency in terms of reduction of required labeled images for fine-tuning in the wildfire scar segmentation task using ViT-base backbone.}
    \label{fig:fire-reduction}
     \end{subfigure}     
     \caption{Evaluation of Prithvi on wildfire scar mapping regarding (a) the performance and (b) the data efficiency of pretrained Prithvi using ViT-base backbone.
     }
     \label{fig:wildfire-results}
\end{figure*}

For the segmentation of wildfire scars, we again observe that the pre-trained model outperforms the same architecture without pretraining in Figure \ref{fig:fire-performance}. The experiments additionally confirm a lower performance when the encoder weights are not updated. This aligns with our expectations, as the decoder head of the model is comparatively small. Finally, the experiments reveal a stable convergence of all models, with the fastest convergence based on the pre-trained weights.

In Figure \ref{fig:fire-reduction}, we confirm that Prithvi allows us to reduce the amount of labeled data significantly while achieving consistent performances. For example, we reduce the number of labeled images during fine-tuning to one forth of the original dataset size (i.e., from 540 images to 252 images to 135 images). The performances are comparable to runs on the full dataset, two-thirds of the dataset, and half of the dataset. This underlines the stability of Prithvi in environments characterized by little labeled data.

We report a comparison of Prithvi to baseline models in Table \ref{tab:fire_scars_results}. Overall, the pretrained version of Prithvi surpasses the version with random weights, as well as U-Net-based and Transformer-based baselines using standard hyperparameters from \cite{mmseg2020}). This model achieves an IoU on the wildfire scar class of 73.62, surpassing a U-Net-based baseline by 2.61pp and a ViT-based baseline by 4.58pp. This improvement is also evident in overall metrics. However, in these metrics, the improvement is less pronounced due to a high class imbalance between the positive class (i.e., wildfire scar) and the negative class (i.e., land). Prithvi profits from pretraining on HLS tiles, improving by up to 1.36pp in IoU on wildfire scar class. 

\begin{table}[ht]
    \centering
    \resizebox{\textwidth}{!}{
    \begin{tabular}{l|cc|ccc}
    \toprule
         & IoU & F1 & mIoU & mF1-score & mAcc  \\
         & (fire scar) & (fire scar) & (both classes) & (both classes) & (both classes)  \\\midrule         
         U-Net (DeepLabV3) \cite{chen2017rethinking} & 71.01 & 83.05  & 83.55 & 90.53 & 87.98  \\
         ViT-base \cite{dosovitskiy2020image} & 69.04 & 81.69 & 82.20 & 89.65 & 90.14 \\[5px] 
         
         Prithvi (not pretrained)  & 72.26 & 83.89 & 84.01 & 90.87 & 92.41 \\
         {Prithvi (pretrained)} & \textbf{73.62} & \textbf{84.81} & \textbf{84.84} & \textbf{91.40} & \textbf{92.48} \\
    \bottomrule
    \end{tabular}
    }
    \caption{Prithvi model performance for the segmentation of wildfire scars compared to a U-Net baseline, as well as recent vision transformer baselines used off-the-shelf with standard hyperparameters from \cite{mmseg2020}. Performance is calculated pixel-wise over the test set, accounting for class imbalance.}
    \label{tab:fire_scars_results}
\end{table}

\subsubsection{Multi-Temporal Crop Segmentation}
The Prithvi 100M model, initially pretrained, is refined further to categorize crops and other land cover types utilizing HLS data and CDL labels from the multi-temporal crop segmentation dataset\footnote{\url{https://huggingface.co/datasets/ibm-nasa-geospatial/multi-temporal-crop-classification}}. To fine-tune the model input chips sized 224x224x18, with dimensions 224 denoting height and width, while 18 results from combining six spectral bands across three time-steps. The bands include Blue, Green, Red, Narrow NIR, SWIR 1, and SWIR 2. The labels, sourced from the dataset, are organized into 13 distinct classes.

For comparison, we used the original U-Net architecture \cite{10.1007/978-3-319-24574-4_28}, characterized by a 32x total downsampling, and trained it on the identical dataset that was utilized for finetuning the GFM. Further, we used flip and rotation augmentations on the training subset. Tversky-focal loss function, combined with normalized class weights, was used for training the model. For optimization purposes, Stochastic Gradient Descent (SGD) with momentum, with sharpness-aware minimization (SAM) framework was used. Additionally, a polynomial learning rate decay policy was applied to ensure a gradually diminishing learning rate, promoting better convergence of the model over time.

\begin{table}[ht]
    \centering
    \small
    \begin{tabular}{l|cc|cccc}
    \toprule
          & \multicolumn{2}{c|}{Prithvi} & \multicolumn{2}{c}{U-Net} \\
         Classes & {Accuracy} & IoU & Accuracy & IoU \\ \midrule
         Natural Vegetation & 46.89\% & 0.4038 & 63.67\% & 0.4578 \\ 
         Forest & 66.38\% & 0.4747 & 71.72\% & 0.4772 \\ 
         Corn & 65.47\% & 0.5491 & 63.33\% & 0.5226 \\ 
         Soybeans & 67.46\% & 0.5297 & 66.77\% & 0.5168 \\ 
         Wetlands & 58.91\% & 0.4020 & 60.36\% & 0.4110 \\ 
         Developed/Barren & 56.49\% & 0.3611 & 60.23\% & 0.4637 \\ 
         Open Water & 90.37\% & 0.6804 & 87.76\% & 0.7596 \\ 
         Winter Wheat & 67.16\% & 0.4967 & 66.39\% & 0.4950 \\ 
         Alfalfa & 66.75\% & 0.3084 & 59.03\% & 0.3848 \\ 
         Fallow/Idle Cropland & 59.23\% & 0.3493 & 52.94\% & 0.3599 \\ 
         Cotton & 66.94\% & 0.3237 & 45.30\% & 0.3258 \\ 
         Sorghum & 73.56\% & 0.3283 & 61.53\% & 0.3910 \\ 
         Other & 47.12\% & 0.3427 & 45.90\% & 0.3268 \\
         \midrule
         Mean & \textbf{64.06}\% & \textbf{0.426} & 61.91\% & \textbf{0.420} \\
    \bottomrule
    \end{tabular}
    \caption{Prithvi model performance for the crop segmentation based on three input timestep compared to a U-Net baseline. For this study, Prithvi was fine-tuned on the CDL dataset for 80 epochs with three input time steps, and U-Net was trained for 100 epochs}
    \label{tab:class_performance_merged}
\end{table}

Table \ref{tab:class_performance_merged} presents a comparative analysis between the Prithvi and U-Net models based on their performance in crop segmentation across various classes. Prithvi exhibits a higher accuracy in classes like Corn, Winter Wheat, Alfalfa, Fallow/Idle Cropland, and Cotton. Prithvi demonstrates a higher IoU in the Corn and Soybeans classes. The mean accuracy for Prithvi of 13 classes is 64.06\%, and for U-Net is 61.91\%. Similarly, the mean IoU for Prithvi is 0.426, and for U-Net is 0.420. Thus, we observe no significant difference between the two models in their IoU scores.

\section{Discussion}

This research paper demonstrates the potential and versatility of employing foundation models within the field of geoscience, with a particular focus on remote sensing applications, which typically necessitate substantial volumes of labeled data. The acquisition of large labeled datasets often engenders issues such as data redundancy or imbalanced distribution of data. Recent works have started to increasingly leverage the concept of self-supervision to remedy this bottleneck. However, to the best of our knowledge, no large-scale model deals with raw satellite imagery, including handling cloud coverage and providing efficient data sampling and loading. To address these challenges, we designed an innovative data preparation pipeline for pre-training purposes. This pipeline encompasses the collation of the HLS dataset for pre-training and the stratification of data according to statistical metrics to facilitate uniform sampling. Key processes such as sampling, ensuring data balance, and maintaining data quality are integral to successfully training a foundation model. Further, in our approach, we incorporated a self-supervised learning mechanism as described in the MAE framework. Within this mechanism, 75\% of the data was masked, prompting the model to attempt to generate the masked information. 
Based on our framework for the generation of geospatial foundation models, we developed a compact foundation model comprising 100 million parameters, which exhibited strong performance across several downstream task categories, notably in segmentation and generation tasks, compared to the state-of-the-art alongside a low sensitivity to the number of labeled images for finetuning. This exemplifies the capability of foundation models to maintain high performance while enabling a broader spectrum of task applications, underpinning the notion that they hold substantial promise in advancing machine learning applications within the geoscience domain.

In the pre-training phase, we made certain assumptions concerning sampling, including a focus on the USA region, a time step of 3, and the utilization of one year's worth of data. Despite exhibiting strong performances in various aspects when compared to other models, our model does not demonstrate a significantly improved performance in handling seasonal changes. Presto represents a self-supervised transformer model for geospatial tasks, which has been pretrained on data from various sensors and particularly accounts for temporal considerations \cite{tseng2023lightweight}. Therefore, we see a complementarity between Presto and our work. The objective of Presto is to facilitate lightweight computations and, accordingly, is comparatively small with less than 1 million parameters, making Presto more than 100 times smaller than Prithvi-100M (see \cite{tseng2023lightweight}). Additionally, the focus of Prithvi is different from Presto. While Prithvi particularly focuses on remote sensing tasks where the predictions are dependent on different surfaces in subregions of the image, Presto focuses on leveraging larger time series of pixels, achieving particularly strong performances in classification and regression tasks. 
Since open-sourcing, Prithvi has already been used by some authors in their studies. Particularly, \cite{li2023assessment} tested Prithvi on flood detection using the Sen1Floods11 dataset and found that Prithvi was outperformed by 2.5\% IoU on the water class by a dedicated U-Net, which is not unexpected as this type of model can very accurately fit the distribution of the floods across known regions (i.e., test set). However, Prithvi strongly outperforms the U-Net on data from unseen geographical regions (i.e., 8.6\% in the IoU on the water class). 
A central promise of foundation models is an improved generalization behavior so that the model is able to adapt rapidly to novel tasks or unseen regions, which is demonstrated by Prithvi in the case of flood mapping. However, we see potential for improvement in the performance of data from the same distribution, which we explore by, e.g., leveraging multi-scale features (see also \cite{li2023assessment}).

By open-sourcing our framework and the resulting foundation model, we hope to contribute to an acceleration in the development of variations of the model to unlock additional downstream tasks accompanied by higher performances and less required labeling effort and computational effort. While our current data pipelines are designed to handle arbitrary data, we have focused on pretraining on data from the U.S. for Prithvi to keep the computational effort manageable. Our experiments have revealed strong generalizability to global data during finetuning; however, we expect an additional increase in performance by pretraining on global HLS data. Although our current model is relatively small, consisting of only 100 million parameters and trained exclusively on data from the US region, it exhibits impressive generalization capabilities across various downstream tasks. The amount of data needed for fine-tuning and generating inferences is notably lower compared to other deep learning algorithms. Nevertheless, in specific downstream tasks, its performance did not surpass state-of-the-art models, prompting us to explore architectural advancements and the possibility of training a larger global model. Our current model architecture is based on a 3D version of the ViT transformer as a backbone. We are eager to explore other backbones, such as 3D Swin transformers, in future work. We have designed our architecture to be able to digest data from additional bands that have not been part of pretraining, which requires further experimentation. Finally, we are interested in the performance of our models on high-resolution satellite imagery (i.e., less than 10m resolution).

\section{Conclusion}

Foundation models can be regarded as versatile tools with both strengths and limitations. It would be unwise to anticipate superior performance across all application categories compared to state-of-the-art models. Nonetheless, they undeniably excel in domains where labeled data is limited. Our framework facilitates the development of large-scale geospatial foundation models, such as Prithvi. Our experiments demonstrate that, based on large-scale, self-supervised pretraining on HLS data, Prithvi is accurate, fast in finetuning, and data-efficient. Importantly, Prithvi generalizes to different resolutions and geo-regions from the entire globe using a few labeled data during fine-tuning. To accelerate work on AI for geoscience and remote sensing, we have open-sourced our code base, model architecture, pretrained weights, fine-tune workflows, and associated demos of the application of the model on downstream tasks. 

\bmhead{Acknowledgments}
We want to express our gratitude to Hugging Face for hosting Prithvi, associated demos, and the corresponding datasets for finetuning. Additionally, we thank the Karlsruhe Service Research and Innovation Hub (KSRI) at Karlsruhe Institute of Technology for supporting PhD studies. 

\clearpage
\bibliography{sn-bibliography_apa}


\begin{thebibliography}{63}
\ifx \bisbn   \undefined \def \bisbn  #1{ISBN #1}\fi
\ifx \binits  \undefined \def \binits#1{#1}\fi
\ifx \bauthor  \undefined \def \bauthor#1{#1}\fi
\ifx \batitle  \undefined \def \batitle#1{#1}\fi
\ifx \bjtitle  \undefined \def \bjtitle#1{#1}\fi
\ifx \bvolume  \undefined \def \bvolume#1{\textbf{#1}}\fi
\ifx \byear  \undefined \def \byear#1{#1}\fi
\ifx \bissue  \undefined \def \bissue#1{#1}\fi
\ifx \bfpage  \undefined \def \bfpage#1{#1}\fi
\ifx \blpage  \undefined \def \blpage #1{#1}\fi
\ifx \burl  \undefined \def \burl#1{\textsf{#1}}\fi
\ifx \doiurl  \undefined \def \doiurl#1{\url{https://doi.org/#1}}\fi
\ifx \betal  \undefined \def \betal{\textit{et al.}}\fi
\ifx \binstitute  \undefined \def \binstitute#1{#1}\fi
\ifx \binstitutionaled  \undefined \def \binstitutionaled#1{#1}\fi
\ifx \bctitle  \undefined \def \bctitle#1{#1}\fi
\ifx \beditor  \undefined \def \beditor#1{#1}\fi
\ifx \bpublisher  \undefined \def \bpublisher#1{#1}\fi
\ifx \bbtitle  \undefined \def \bbtitle#1{#1}\fi
\ifx \bedition  \undefined \def \bedition#1{#1}\fi
\ifx \bseriesno  \undefined \def \bseriesno#1{#1}\fi
\ifx \blocation  \undefined \def \blocation#1{#1}\fi
\ifx \bsertitle  \undefined \def \bsertitle#1{#1}\fi
\ifx \bsnm \undefined \def \bsnm#1{#1}\fi
\ifx \bsuffix \undefined \def \bsuffix#1{#1}\fi
\ifx \bparticle \undefined \def \bparticle#1{#1}\fi
\ifx \barticle \undefined \def \barticle#1{#1}\fi
\bibcommenthead
\ifx \bconfdate \undefined \def \bconfdate #1{#1}\fi
\ifx \botherref \undefined \def \botherref #1{#1}\fi
\ifx \url \undefined \def \url#1{\textsf{#1}}\fi
\ifx \bchapter \undefined \def \bchapter#1{#1}\fi
\ifx \bbook \undefined \def \bbook#1{#1}\fi
\ifx \bcomment \undefined \def \bcomment#1{#1}\fi
\ifx \oauthor \undefined \def \oauthor#1{#1}\fi
\ifx \citeauthoryear \undefined \def \citeauthoryear#1{#1}\fi
\ifx \endbibitem  \undefined \def \endbibitem {}\fi
\ifx \bconflocation  \undefined \def \bconflocation#1{#1}\fi
\ifx \arxivurl  \undefined \def \arxivurl#1{\textsf{#1}}\fi
\csname PreBibitemsHook\endcsname

%%% 1
\bibitem[\protect\citeauthoryear{Alemohammad et~al.}{2020}]{Hamed2020Advancing}
\begin{botherref}
\oauthor{\bsnm{Alemohammad}, \binits{H.}},
\oauthor{\bsnm{Maskey}, \binits{M.}},
\oauthor{\bsnm{Estes}, \binits{L.}},
\oauthor{\bsnm{Gentine}, \binits{P.}},
\oauthor{\bsnm{Lunga}, \binits{D.}},
\oauthor{\bsnm{Fang}, \binits{Z.}}:
{Advancing Application of Machine Learning Tools for NASA’s Earth Observation
  Data }
(2020).
\url{https://www.earthdata.nasa.gov/s3fs-public/imported/NASA_ML_Workshop_Report.pdf}
\end{botherref}
\endbibitem

%%% 2
\bibitem[\protect\citeauthoryear{Touvron
  et~al.}{2023}]{10.48550/arxiv.2302.13971}
\begin{botherref}
\oauthor{\bsnm{Touvron}, \binits{H.}},
\oauthor{\bsnm{Lavril}, \binits{T.}},
\oauthor{\bsnm{Izacard}, \binits{G.}},
\oauthor{\bsnm{Martinet}, \binits{X.}},
\oauthor{\bsnm{Lachaux}, \binits{M.-A.}},
\oauthor{\bsnm{Lacroix}, \binits{T.}},
\oauthor{\bsnm{Rozière}, \binits{B.}},
\oauthor{\bsnm{Goyal}, \binits{N.}},
\oauthor{\bsnm{Hambro}, \binits{E.}},
\oauthor{\bsnm{Azhar}, \binits{F.}},
\oauthor{\bsnm{Rodriguez}, \binits{A.}},
\oauthor{\bsnm{Joulin}, \binits{A.}},
\oauthor{\bsnm{Grave}, \binits{E.}},
\oauthor{\bsnm{Lample}, \binits{G.}}:
{{LLaMA: Open and Efficient Foundation Language Models}}.
{Preprint Available on arXiv:2302.13971}
(2023)
\end{botherref}
\endbibitem

%%% 3
\bibitem[\protect\citeauthoryear{Yuan et~al.}{2021}]{10.48550/arxiv.2111.11432}
\begin{botherref}
\oauthor{\bsnm{Yuan}, \binits{L.}},
\oauthor{\bsnm{Chen}, \binits{D.}},
\oauthor{\bsnm{Chen}, \binits{Y.-L.}},
\oauthor{\bsnm{Codella}, \binits{N.}},
\oauthor{\bsnm{Dai}, \binits{X.}},
\oauthor{\bsnm{Gao}, \binits{J.}},
\oauthor{\bsnm{Hu}, \binits{H.}},
\oauthor{\bsnm{Huang}, \binits{X.}},
\oauthor{\bsnm{Li}, \binits{B.}},
\oauthor{\bsnm{Li}, \binits{C.}},
\oauthor{\bsnm{Liu}, \binits{C.}},
\oauthor{\bsnm{Liu}, \binits{M.}},
\oauthor{\bsnm{Liu}, \binits{Z.}},
\oauthor{\bsnm{Lu}, \binits{Y.}},
\oauthor{\bsnm{Shi}, \binits{Y.}},
\oauthor{\bsnm{Wang}, \binits{L.}},
\oauthor{\bsnm{Wang}, \binits{J.}},
\oauthor{\bsnm{Xiao}, \binits{B.}},
\oauthor{\bsnm{Xiao}, \binits{Z.}},
\oauthor{\bsnm{Yang}, \binits{J.}},
\oauthor{\bsnm{Zeng}, \binits{M.}},
\oauthor{\bsnm{Zhou}, \binits{L.}},
\oauthor{\bsnm{Zhang}, \binits{P.}}:
{{Florence: A New Foundation Model for Computer Vision}}.
{Preprint Available on arXiv:2111.11432}
(2021)
\end{botherref}
\endbibitem

%%% 4
\bibitem[\protect\citeauthoryear{Liu et~al.}{2023}]{10.48550/arxiv.2306.11029}
\begin{botherref}
\oauthor{\bsnm{Liu}, \binits{F.}},
\oauthor{\bsnm{Chen}, \binits{D.}},
\oauthor{\bsnm{Guan}, \binits{Z.}},
\oauthor{\bsnm{Zhou}, \binits{X.}},
\oauthor{\bsnm{Zhu}, \binits{J.}},
\oauthor{\bsnm{Zhou}, \binits{J.}}:
{{RemoteCLIP: A Vision Language Foundation Model for Remote Sensing}}.
{Preprint Available on arXiv:2306.11029}
(2023)
\end{botherref}
\endbibitem

%%% 5
\bibitem[\protect\citeauthoryear{Yu et~al.}{2022}]{10.48550/arxiv.2205.01917}
\begin{botherref}
\oauthor{\bsnm{Yu}, \binits{J.}},
\oauthor{\bsnm{Wang}, \binits{Z.}},
\oauthor{\bsnm{Vasudevan}, \binits{V.}},
\oauthor{\bsnm{Yeung}, \binits{L.}},
\oauthor{\bsnm{Seyedhosseini}, \binits{M.}},
\oauthor{\bsnm{Wu}, \binits{Y.}}:
{{CoCa: Contrastive Captioners Are Image-Text Foundation Models}}.
{{Transactions on Machine Learning Research}}
(2022)
\end{botherref}
\endbibitem

%%% 6
\bibitem[\protect\citeauthoryear{Wang et~al.}{2022}]{10.48550/arxiv.2209.07526}
\begin{barticle}
\bauthor{\bsnm{Wang}, \binits{J.}},
\bauthor{\bsnm{Chen}, \binits{D.}},
\bauthor{\bsnm{Wu}, \binits{Z.}},
\bauthor{\bsnm{Luo}, \binits{C.}},
\bauthor{\bsnm{Zhou}, \binits{L.}},
\bauthor{\bsnm{Zhao}, \binits{Y.}},
\bauthor{\bsnm{Xie}, \binits{Y.}},
\bauthor{\bsnm{Liu}, \binits{C.}},
\bauthor{\bsnm{Jiang}, \binits{Y.-G.}},
\bauthor{\bsnm{Yuan}, \binits{L.}}:
\batitle{{{OmniVL: One Foundation Model for Image-Language and Video-Language
  Tasks}}}.
\bjtitle{{Advances in Neural Information Processing Systems}}
\bvolume{35},
\bfpage{5696}--\blpage{5710}
(\byear{2022})
\end{barticle}
\endbibitem

%%% 7
\bibitem[\protect\citeauthoryear{Cong et~al.}{2022}]{satmae}
\begin{barticle}
\bauthor{\bsnm{Cong}, \binits{Y.}},
\bauthor{\bsnm{Khanna}, \binits{S.}},
\bauthor{\bsnm{Meng}, \binits{C.}},
\bauthor{\bsnm{Liu}, \binits{P.}},
\bauthor{\bsnm{Rozi}, \binits{E.}},
\bauthor{\bsnm{He}, \binits{Y.}},
\bauthor{\bsnm{Burke}, \binits{M.}},
\bauthor{\bsnm{Lobell}, \binits{D.}},
\bauthor{\bsnm{Ermon}, \binits{S.}}:
\batitle{{Satmae: Pre-Training Transformers for Temporal and Multi-Spectral
  Satellite Imagery}}.
\bjtitle{{Advances in Neural Information Processing Systems}}
\bvolume{35},
\bfpage{197}--\blpage{211}
(\byear{2022})
\end{barticle}
\endbibitem

%%% 8
\bibitem[\protect\citeauthoryear{Xia et~al.}{2018}]{10.48550/arxiv.1711.10398}
\begin{bchapter}
\bauthor{\bsnm{Xia}, \binits{G.-S.}},
\bauthor{\bsnm{Bai}, \binits{X.}},
\bauthor{\bsnm{Ding}, \binits{J.}},
\bauthor{\bsnm{Zhu}, \binits{Z.}},
\bauthor{\bsnm{Belongie}, \binits{S.}},
\bauthor{\bsnm{Luo}, \binits{J.}},
\bauthor{\bsnm{Datcu}, \binits{M.}},
\bauthor{\bsnm{Pelillo}, \binits{M.}},
\bauthor{\bsnm{Zhang}, \binits{L.}}:
\bctitle{{{DOTA: A Large-Scale Dataset for Object Detection in Aerial
  Images}}}.
In: \bbtitle{The IEEE Conference on Computer Vision and Pattern Recognition
  (CVPR)},
pp. \bfpage{3974}--\blpage{3983}
(\byear{2018})
\end{bchapter}
\endbibitem

%%% 9
\bibitem[\protect\citeauthoryear{Sun et~al.}{2023}]{10.1109/tgrs.2022.3194732}
\begin{barticle}
\bauthor{\bsnm{Sun}, \binits{X.}},
\bauthor{\bsnm{Wang}, \binits{P.}},
\bauthor{\bsnm{Lu}, \binits{W.}},
\bauthor{\bsnm{Zhu}, \binits{Z.}},
\bauthor{\bsnm{Lu}, \binits{X.}},
\bauthor{\bsnm{He}, \binits{Q.}},
\bauthor{\bsnm{Li}, \binits{J.}},
\bauthor{\bsnm{Rong}, \binits{X.}},
\bauthor{\bsnm{Yang}, \binits{Z.}},
\bauthor{\bsnm{Chang}, \binits{H.}},
\bauthor{\bsnm{He}, \binits{Q.}},
\bauthor{\bsnm{Yang}, \binits{G.}},
\bauthor{\bsnm{Wang}, \binits{R.}},
\bauthor{\bsnm{Lu}, \binits{J.}},
\bauthor{\bsnm{Fu}, \binits{K.}}:
\batitle{{{RingMo: A Remote Sensing Foundation Model With Masked Image
  Modeling}}}.
\bjtitle{{IEEE Transactions on Geoscience and Remote Sensing}}
\bvolume{61},
\bfpage{1}--\blpage{22}
(\byear{2023})
\end{barticle}
\endbibitem

%%% 10
\bibitem[\protect\citeauthoryear{Cha et~al.}{2023}]{10.48550/arxiv.2304.05215}
\begin{botherref}
\oauthor{\bsnm{Cha}, \binits{K.}},
\oauthor{\bsnm{Seo}, \binits{J.}},
\oauthor{\bsnm{Lee}, \binits{T.}}:
{{A Billion-Scale Foundation Model for Remote Sensing Images}}.
Preprint available on arXiv:2304.05215
(2023)
\end{botherref}
\endbibitem

%%% 11
\bibitem[\protect\citeauthoryear{Nguyen et~al.}{2023}]{nguyen2023climax}
\begin{botherref}
\oauthor{\bsnm{Nguyen}, \binits{T.}},
\oauthor{\bsnm{Brandstetter}, \binits{J.}},
\oauthor{\bsnm{Kapoor}, \binits{A.}},
\oauthor{\bsnm{Gupta}, \binits{J.K.}},
\oauthor{\bsnm{Grover}, \binits{A.}}:
{ClimaX: A Foundation Model for Weather and Climate}.
Preprint available on arXiv:2301.10343
(2023)
\end{botherref}
\endbibitem

%%% 12
\bibitem[\protect\citeauthoryear{Mukkavilli et~al.}{2023}]{mukkavilli2023ai}
\begin{botherref}
\oauthor{\bsnm{Mukkavilli}, \binits{S.K.}},
\oauthor{\bsnm{Civitarese}, \binits{D.S.}},
\oauthor{\bsnm{Schmude}, \binits{J.}},
\oauthor{\bsnm{Jakubik}, \binits{J.}},
\oauthor{\bsnm{Jones}, \binits{A.}},
\oauthor{\bsnm{Nguyen}, \binits{N.}},
\oauthor{\bsnm{Phillips}, \binits{C.}},
\oauthor{\bsnm{Roy}, \binits{S.}},
\oauthor{\bsnm{Singh}, \binits{S.}},
\oauthor{\bsnm{Watson}, \binits{C.}}, et al.:
{AI Foundation Models for Weather and Climate: Applications, Design, and
  Implementation}.
{Preprint Available on arXiv:2309.10808}
(2023)
\end{botherref}
\endbibitem

%%% 13
\bibitem[\protect\citeauthoryear{Ma et~al.}{2019}]{ma2019deep}
\begin{barticle}
\bauthor{\bsnm{Ma}, \binits{L.}},
\bauthor{\bsnm{Liu}, \binits{Y.}},
\bauthor{\bsnm{Zhang}, \binits{X.}},
\bauthor{\bsnm{Ye}, \binits{Y.}},
\bauthor{\bsnm{Yin}, \binits{G.}},
\bauthor{\bsnm{Johnson}, \binits{B.A.}}:
\batitle{{Deep Learning in Remote Sensing Applications: A Meta-Analysis and
  Review}}.
\bjtitle{{ISPRS Journal of Photogrammetry and Remote Sensing}}
\bvolume{152},
\bfpage{166}--\blpage{177}
(\byear{2019})
\end{barticle}
\endbibitem

%%% 14
\bibitem[\protect\citeauthoryear{Aleissaee
  et~al.}{2023}]{aleissaee2023transformers}
\begin{barticle}
\bauthor{\bsnm{Aleissaee}, \binits{A.A.}},
\bauthor{\bsnm{Kumar}, \binits{A.}},
\bauthor{\bsnm{Anwer}, \binits{R.M.}},
\bauthor{\bsnm{Khan}, \binits{S.}},
\bauthor{\bsnm{Cholakkal}, \binits{H.}},
\bauthor{\bsnm{Xia}, \binits{G.-S.}},
\bauthor{\bsnm{Khan}, \binits{F.S.}}:
\batitle{{Transformers in Remote Sensing: A Survey}}.
\bjtitle{{Remote Sensing}}
\bvolume{15}(\bissue{7}),
\bfpage{1860}
(\byear{2023})
\end{barticle}
\endbibitem

%%% 15
\bibitem[\protect\citeauthoryear{Ronneberger et~al.}{2015}]{ronneberger2015u}
\begin{bchapter}
\bauthor{\bsnm{Ronneberger}, \binits{O.}},
\bauthor{\bsnm{Fischer}, \binits{P.}},
\bauthor{\bsnm{Brox}, \binits{T.}}:
\bctitle{{{"U-Net: Convolutional Networks for Biomedical Image
  Segmentation"}}}.
In: \bbtitle{Proceedings of the 18th International Conference on Medical Image
  Computing and Computer-Assisted Intervention (MICCAI)},
pp. \bfpage{234}--\blpage{241}
(\byear{2015})
\end{bchapter}
\endbibitem

%%% 16
\bibitem[\protect\citeauthoryear{Wang et~al.}{2022}]{10.1155/2022/1603273}
\begin{barticle}
\bauthor{\bsnm{Wang}, \binits{Y.}},
\bauthor{\bsnm{Kong}, \binits{J.}},
\bauthor{\bsnm{Zhang}, \binits{H.}}:
\batitle{{{U-Net: A Smart Application With Multidimensional Attention Network
  for Remote Sensing Images}}}.
\bjtitle{{Scientific Programming}}
\bvolume{2022},
\bfpage{1}--\blpage{11}
(\byear{2022})
\end{barticle}
\endbibitem

%%% 17
\bibitem[\protect\citeauthoryear{Qin et~al.}{2022}]{10.1109/lgrs.2020.3047918}
\begin{barticle}
\bauthor{\bsnm{Qin}, \binits{P.}},
\bauthor{\bsnm{Cai}, \binits{Y.}},
\bauthor{\bsnm{Wang}, \binits{X.}}:
\batitle{{{Small Waterbody Extraction With Improved U-Net Using Zhuhai-1
  Hyperspectral Remote Sensing Images}}}.
\bjtitle{{IEEE Geoscience and Remote Sensing Letters}}
\bvolume{19},
\bfpage{1}--\blpage{5}
(\byear{2022})
\end{barticle}
\endbibitem

%%% 18
\bibitem[\protect\citeauthoryear{He et~al.}{2022}]{he2022masked}
\begin{bchapter}
\bauthor{\bsnm{He}, \binits{K.}},
\bauthor{\bsnm{Chen}, \binits{X.}},
\bauthor{\bsnm{Xie}, \binits{S.}},
\bauthor{\bsnm{Li}, \binits{Y.}},
\bauthor{\bsnm{Doll{\'a}r}, \binits{P.}},
\bauthor{\bsnm{Girshick}, \binits{R.}}:
\bctitle{{Masked Autoencoders Are Scalable Vision Learners}}.
In: \bbtitle{Proceedings of the IEEE/CVF Conference on Computer Vision and
  Pattern Recognition},
pp. \bfpage{16000}--\blpage{16009}
(\byear{2022})
\end{bchapter}
\endbibitem

%%% 19
\bibitem[\protect\citeauthoryear{Dosovitskiy
  et~al.}{2021}]{dosovitskiy2020image}
\begin{bchapter}
\bauthor{\bsnm{Dosovitskiy}, \binits{A.}},
\bauthor{\bsnm{Beyer}, \binits{L.}},
\bauthor{\bsnm{Kolesnikov}, \binits{A.}},
\bauthor{\bsnm{Weissenborn}, \binits{D.}},
\bauthor{\bsnm{Zhai}, \binits{X.}},
\bauthor{\bsnm{Unterthiner}, \binits{T.}},
\bauthor{\bsnm{Dehghani}, \binits{M.}},
\bauthor{\bsnm{Minderer}, \binits{M.}},
\bauthor{\bsnm{Heigold}, \binits{G.}},
\bauthor{\bsnm{Gelly}, \binits{S.}}, \betal:
\bctitle{{An Image Is Worth 16x16 Words: Transformers for Image Recognition at
  Scale}}.
In: \bbtitle{Proceedings of the International Conference on Learning
  Representations}
(\byear{2021})
\end{bchapter}
\endbibitem

%%% 20
\bibitem[\protect\citeauthoryear{Jakubik et~al.}{2023}]{jakubik2023toward}
\begin{bchapter}
\bauthor{\bsnm{Jakubik}, \binits{J.}},
\bauthor{\bsnm{Muszynski}, \binits{M.}},
\bauthor{\bsnm{V{\"o}ssing}, \binits{M.}},
\bauthor{\bsnm{K{\"u}hl}, \binits{N.}},
\bauthor{\bsnm{Brunschwiler}, \binits{T.}}:
\bctitle{{Toward Foundation Models for Earth Monitoring: Generalizable Deep
  Learning Models for Natural Hazard Segmentation}}.
In: \bbtitle{Proceedings of the International Geoscience and Remote Sensing
  Symposium (IGARSS)},
pp. \bfpage{5638}--\blpage{5641}
(\byear{2023})
\end{bchapter}
\endbibitem

%%% 21
\bibitem[\protect\citeauthoryear{Tseng et~al.}{2023}]{tseng2023lightweight}
\begin{botherref}
\oauthor{\bsnm{Tseng}, \binits{G.}},
\oauthor{\bsnm{Zvonkov}, \binits{I.}},
\oauthor{\bsnm{Purohit}, \binits{M.}},
\oauthor{\bsnm{Rolnick}, \binits{D.}},
\oauthor{\bsnm{Kerner}, \binits{H.}}:
{Lightweight, Pre-Trained Transformers for Remote Sensing Timeseries}.
{Preprint Available on arXiv:2304.14065}
(2023)
\end{botherref}
\endbibitem

%%% 22
\bibitem[\protect\citeauthoryear{Deshpande
  et~al.}{2022}]{10.1016/j.mex.2022.101741}
\begin{barticle}
\bauthor{\bsnm{Deshpande}, \binits{M.V.}},
\bauthor{\bsnm{Pillai}, \binits{D.}},
\bauthor{\bsnm{Jain}, \binits{M.}}:
\batitle{{{Agricultural Burned Area Detection Using an Integrated Approach
  Utilizing Multi Spectral Instrument Based Fire and Vegetation Indices From
  Sentinel-2 Satellite}}}.
\bjtitle{{MethodsX}}
\bvolume{9},
\bfpage{101741}
(\byear{2022})
\end{barticle}
\endbibitem

%%% 23
\bibitem[\protect\citeauthoryear{Bar
  et~al.}{2020}]{10.1016/j.rsase.2020.100324}
\begin{barticle}
\bauthor{\bsnm{Bar}, \binits{S.}},
\bauthor{\bsnm{Parida}, \binits{B.R.}},
\bauthor{\bsnm{Pandey}, \binits{A.C.}}:
\batitle{{{Landsat-8 and Sentinel-2 Based Forest Fire Burn Area Mapping Using
  Machine Learning Algorithms on GEE Cloud Platform Over Uttarakhand, Western
  Himalaya}}}.
\bjtitle{{Remote Sensing Applications: Society and Environment}}
\bvolume{18},
\bfpage{100324}
(\byear{2020})
\end{barticle}
\endbibitem

%%% 24
\bibitem[\protect\citeauthoryear{Brovkina
  et~al.}{2020}]{10.1080/19475705.2020.1836037}
\begin{barticle}
\bauthor{\bsnm{Brovkina}, \binits{O.}},
\bauthor{\bsnm{Stojanović}, \binits{M.}},
\bauthor{\bsnm{Milanović}, \binits{S.}},
\bauthor{\bsnm{Latypov}, \binits{I.}},
\bauthor{\bsnm{Marković}, \binits{N.}},
\bauthor{\bsnm{Cienciala}, \binits{E.}}:
\batitle{{{Monitoring of Post-Fire Forest Scars in Serbia Based on Satellite
  Sentinel-2 Data}}}.
\bjtitle{{Geomatics, Natural Hazards and Risk}}
\bvolume{11}(\bissue{1}),
\bfpage{2315}--\blpage{2339}
(\byear{2020})
\end{barticle}
\endbibitem

%%% 25
\bibitem[\protect\citeauthoryear{Dimitris
  et~al.}{2020}]{10.4236/jgis.2020.123014}
\begin{barticle}
\bauthor{\bsnm{Dimitris}, \binits{S.}},
\bauthor{\bsnm{Thomas}, \binits{K.}},
\bauthor{\bsnm{Chara}, \binits{M.}},
\bauthor{\bsnm{Ioannis}, \binits{Z.G.}}:
\batitle{{{Automated Burned Scar Mapping Using Sentinel-2 Imagery}}}.
\bjtitle{{Journal of Geographic Information System}}
\bvolume{12}(\bissue{03}),
\bfpage{221}--\blpage{240}
(\byear{2020})
\end{barticle}
\endbibitem

%%% 26
\bibitem[\protect\citeauthoryear{Hu et~al.}{2021}]{bs_unet1}
\begin{barticle}
\bauthor{\bsnm{Hu}, \binits{X.}},
\bauthor{\bsnm{Ban}, \binits{Y.}},
\bauthor{\bsnm{Nascetti}, \binits{A.}}:
\batitle{{{Uni-Temporal Multispectral Imagery for Burned Area Mapping With Deep
  Learning}}}.
\bjtitle{{Remote Sensing}}
\bvolume{13}(\bissue{8}),
\bfpage{1509}
(\byear{2021})
\end{barticle}
\endbibitem

%%% 27
\bibitem[\protect\citeauthoryear{Zhang et~al.}{2021}]{bs_unet2}
\begin{barticle}
\bauthor{\bsnm{Zhang}, \binits{P.}},
\bauthor{\bsnm{Ban}, \binits{Y.}},
\bauthor{\bsnm{Nascetti}, \binits{A.}}:
\batitle{{{Learning U-Net Without Forgetting for Near Real-Time Wildfire
  Monitoring by the Fusion of SAR and Optical Time Series}}}.
\bjtitle{{Remote Sensing of Environment}}
\bvolume{261},
\bfpage{112467}
(\byear{2021})
\end{barticle}
\endbibitem

%%% 28
\bibitem[\protect\citeauthoryear{Arruda et~al.}{2021}]{bs_dnn}
\begin{barticle}
\bauthor{\bsnm{Arruda}, \binits{V.L.S.}},
\bauthor{\bsnm{Piontekowski}, \binits{V.J.}},
\bauthor{\bsnm{Alencar}, \binits{A.}},
\bauthor{\bsnm{Pereira}, \binits{R.S.}},
\bauthor{\bsnm{Matricardi}, \binits{E.A.T.}}:
\batitle{{{An Alternative Approach for Mapping Burn Scars Using Landsat
  Imagery, Google Earth Engine, and Deep Learning in the Brazilian Savanna}}}.
\bjtitle{{Remote Sensing Applications: Society and Environment}}
\bvolume{22},
\bfpage{100472}
(\byear{2021})
\end{barticle}
\endbibitem

%%% 29
\bibitem[\protect\citeauthoryear{Sahoo et~al.}{2006}]{EarlyFloodDL0}
\begin{barticle}
\bauthor{\bsnm{Sahoo}, \binits{G.B.}},
\bauthor{\bsnm{Ray}, \binits{C.}},
\bauthor{\bsnm{Carlo}, \binits{E.H.D.}}:
\batitle{{{Use of Neural Network to Predict Flash Flood and Attendant Water
  Qualities of a Mountainous Stream on Oahu, Hawaii}}}.
\bjtitle{{Journal of Hydrology}}
\bvolume{327}(\bissue{3-4}),
\bfpage{525}--\blpage{538}
(\byear{2006})
\end{barticle}
\endbibitem

%%% 30
\bibitem[\protect\citeauthoryear{Yang et~al.}{2015}]{EarlyFloodDL1}
\begin{barticle}
\bauthor{\bsnm{Yang}, \binits{L.}},
\bauthor{\bsnm{Tian}, \binits{S.}},
\bauthor{\bsnm{Yu}, \binits{L.}},
\bauthor{\bsnm{Ye}, \binits{F.}},
\bauthor{\bsnm{Qian}, \binits{J.}},
\bauthor{\bsnm{Qian}, \binits{Y.}}:
\batitle{{{Deep Learning for Extracting Water Body From Landsat Imagery}}}.
\bjtitle{{International Journal of Innovative Computing, Information and
  Control}}
\bvolume{11}(\bissue{06}),
\bfpage{1913}
(\byear{2015})
\end{barticle}
\endbibitem

%%% 31
\bibitem[\protect\citeauthoryear{Gong et~al.}{2016}]{EarlyFloodDL2}
\begin{barticle}
\bauthor{\bsnm{Gong}, \binits{M.}},
\bauthor{\bsnm{Zhao}, \binits{J.}},
\bauthor{\bsnm{Liu}, \binits{J.}},
\bauthor{\bsnm{Miao}, \binits{Q.}},
\bauthor{\bsnm{Jiao}, \binits{L.}}:
\batitle{{{Change Detection in Synthetic Aperture Radar Images Based on Deep
  Neural Networks}}}.
\bjtitle{{IEEE Transactions on Neural Networks and Learning Systems}}
\bvolume{27}(\bissue{1}),
\bfpage{125}--\blpage{138}
(\byear{2016})
\end{barticle}
\endbibitem

%%% 32
\bibitem[\protect\citeauthoryear{Ghosh et~al.}{2022}]{Flood_SAR}
\begin{barticle}
\bauthor{\bsnm{Ghosh}, \binits{B.}},
\bauthor{\bsnm{Garg}, \binits{S.}},
\bauthor{\bsnm{Motagh}, \binits{M.}}:
\batitle{{{Automatic Flood Detection From Sentinel-1 Data Using Deep Learning
  Architectures}}}.
\bjtitle{{ISPRS Annals of the Photogrammetry, Remote Sensing and Spatial
  Information Sciences}}
\bvolume{3},
\bfpage{201}--\blpage{208}
(\byear{2022})
\end{barticle}
\endbibitem

%%% 33
\bibitem[\protect\citeauthoryear{Tuyen et~al.}{2021}]{Flood_PsoUnet}
\begin{barticle}
\bauthor{\bsnm{Tuyen}, \binits{D.N.}},
\bauthor{\bsnm{Tuan}, \binits{T.M.}},
\bauthor{\bsnm{Son}, \binits{L.H.}},
\bauthor{\bsnm{Ngan}, \binits{T.T.}},
\bauthor{\bsnm{Giang}, \binits{N.L.}},
\bauthor{\bsnm{Thong}, \binits{P.H.}},
\bauthor{\bsnm{Hieu}, \binits{V.V.}},
\bauthor{\bsnm{Gerogiannis}, \binits{V.C.}},
\bauthor{\bsnm{Tzimos}, \binits{D.}},
\bauthor{\bsnm{Kanavos}, \binits{A.}}:
\batitle{{{A Novel Approach Combining Particle Swarm Optimization and Deep
  Learning for Flash Flood Detection From Satellite Images}}}.
\bjtitle{{Mathematics}}
\bvolume{9}(\bissue{22}),
\bfpage{2846}
(\byear{2021})
\end{barticle}
\endbibitem

%%% 34
\bibitem[\protect\citeauthoryear{Zhang and Xia}{2021}]{Flood_OpticalSAR}
\begin{barticle}
\bauthor{\bsnm{Zhang}, \binits{L.}},
\bauthor{\bsnm{Xia}, \binits{J.}}:
\batitle{{{Flood Detection Using Multiple Chinese Satellite Datasets During
  2020 China Summer Floods}}}.
\bjtitle{{Remote Sensing}}
\bvolume{14}(\bissue{1}),
\bfpage{51}
(\byear{2021})
\end{barticle}
\endbibitem

%%% 35
\bibitem[\protect\citeauthoryear{Du et~al.}{2019}]{CropTypeDSS}
\begin{barticle}
\bauthor{\bsnm{Du}, \binits{Z.}},
\bauthor{\bsnm{Yang}, \binits{J.}},
\bauthor{\bsnm{Ou}, \binits{C.}},
\bauthor{\bsnm{Zhang}, \binits{T.}}:
\batitle{{{Smallholder Crop Area Mapped With a Semantic Segmentation Deep
  Learning Method}}}.
\bjtitle{{Remote Sensing}}
\bvolume{11}(\bissue{7}),
\bfpage{888}
(\byear{2019})
\end{barticle}
\endbibitem

%%% 36
\bibitem[\protect\citeauthoryear{Wang et~al.}{2022}]{CropTypeUNet}
\begin{barticle}
\bauthor{\bsnm{Wang}, \binits{L.}},
\bauthor{\bsnm{Wang}, \binits{J.}},
\bauthor{\bsnm{Zhang}, \binits{X.}},
\bauthor{\bsnm{Wang}, \binits{L.}},
\bauthor{\bsnm{Qin}, \binits{F.}}:
\batitle{{{Deep Segmentation and Classification of Complex Crops Using
  Multi-Feature Satellite Imagery}}}.
\bjtitle{{Computers and Electronics in Agriculture}}
\bvolume{200},
\bfpage{107249}
(\byear{2022})
\end{barticle}
\endbibitem

%%% 37
\bibitem[\protect\citeauthoryear{Mohammadi et~al.}{2023}]{CropTypeCNN}
\begin{barticle}
\bauthor{\bsnm{Mohammadi}, \binits{S.}},
\bauthor{\bsnm{Belgiu}, \binits{M.}},
\bauthor{\bsnm{Stein}, \binits{A.}}:
\batitle{{{Improvement in Crop Mapping From Satellite Image Time Series by
  Effectively Supervising Deep Neural Networks}}}.
\bjtitle{{ISPRS Journal of Photogrammetry and Remote Sensing}}
\bvolume{198},
\bfpage{272}--\blpage{283}
(\byear{2023})
\end{barticle}
\endbibitem

%%% 38
\bibitem[\protect\citeauthoryear{Khan et~al.}{2023}]{CropTypeSS}
\begin{bchapter}
\bauthor{\bsnm{Khan}, \binits{A.H.}},
\bauthor{\bsnm{Zafar}, \binits{Z.}},
\bauthor{\bsnm{Shahzad}, \binits{M.}},
\bauthor{\bsnm{Berns}, \binits{K.}},
\bauthor{\bsnm{Fraz}, \binits{M.M.}}:
\bctitle{{{Crop Type Classification Using Multi-Temporal Sentinel-2 Satellite
  Imagery: A Deep Semantic Segmentation Approach}}}.
In: \bbtitle{Proceedings of the International Conference on Robotics and
  Automation in Industry (ICRAI)},
pp. \bfpage{1}--\blpage{6}
(\byear{2023})
\end{bchapter}
\endbibitem

%%% 39
\bibitem[\protect\citeauthoryear{Claverie et~al.}{2018}]{HLSpub}
\begin{barticle}
\bauthor{\bsnm{Claverie}, \binits{M.}},
\bauthor{\bsnm{Ju}, \binits{J.}},
\bauthor{\bsnm{Masek}, \binits{J.G.}},
\bauthor{\bsnm{Dungan}, \binits{J.L.}},
\bauthor{\bsnm{Vermote}, \binits{E.F.}},
\bauthor{\bsnm{Roger}, \binits{J.-C.}},
\bauthor{\bsnm{Skakun}, \binits{S.V.}},
\bauthor{\bsnm{Justice}, \binits{C.}}:
\batitle{{{The Harmonized Landsat and Sentinel-2 Surface Reflectance Data
  Set}}}.
\bjtitle{{Remote Sensing of Environment}}
\bvolume{219},
\bfpage{145}--\blpage{161}
(\byear{2018})
\end{barticle}
\endbibitem

%%% 40
\bibitem[\protect\citeauthoryear{Masek et~al.}{2021a}]{HLSL30}
\begin{botherref}
\oauthor{\bsnm{Masek}, \binits{J.}},
\oauthor{\bsnm{Ju}, \binits{J.}},
\oauthor{\bsnm{Roger}, \binits{J.}},
\oauthor{\bsnm{Skakun}, \binits{S.}},
\oauthor{\bsnm{Vermote}, \binits{E.}},
\oauthor{\bsnm{Claverie}, \binits{M.}},
\oauthor{\bsnm{Dungan}, \binits{J.}},
\oauthor{\bsnm{Yin}, \binits{Z.}},
\oauthor{\bsnm{Freitag}, \binits{B.}},
\oauthor{\bsnm{Justice}, \binits{C.}}:
{{HLS Operational Land Imager Surface Reflectance and TOA Brightness Daily
  Global 30 M v2.0}}.
NASA EOSDIS Land Processes DAAC
(2021)
\end{botherref}
\endbibitem

%%% 41
\bibitem[\protect\citeauthoryear{Masek et~al.}{2021b}]{HLSS30}
\begin{botherref}
\oauthor{\bsnm{Masek}, \binits{J.}},
\oauthor{\bsnm{Ju}, \binits{J.}},
\oauthor{\bsnm{Roger}, \binits{J.}},
\oauthor{\bsnm{Skakun}, \binits{S.}},
\oauthor{\bsnm{Vermote}, \binits{E.}},
\oauthor{\bsnm{Claverie}, \binits{M.}},
\oauthor{\bsnm{Dungan}, \binits{J.}},
\oauthor{\bsnm{Yin}, \binits{Z.}},
\oauthor{\bsnm{Freitag}, \binits{B.}},
\oauthor{\bsnm{Justice}, \binits{C.}}:
{{HLS Sentinel-2 MSI Surface Reflectance Daily Global 30m v2.0}}.
NASA EOSDIS Land Processes DAAC
(2021)
\end{botherref}
\endbibitem

%%% 42
\bibitem[\protect\citeauthoryear{Krehbiel and Jami}{}]{hls_super}
\begin{botherref}
\oauthor{\bsnm{Krehbiel}, \binits{C.}},
\oauthor{\bsnm{Jami}, \binits{M.}}:
HLS Subsetting, Processing, and Exporting Reformatted Data Prep Script.
\url{https://git.earthdata.nasa.gov/projects/LPDUR/repos/hls-super-script/browse}
\end{botherref}
\endbibitem

%%% 43
\bibitem[\protect\citeauthoryear{NASA}{}]{LPDAAC}
\begin{botherref}
\oauthor{\bsnm{NASA}}:
{HLS Overview}.
\url{https://lpdaac.usgs.gov/data/get-started-data/collection-overview/missions/harmonized-landsat-sentinel-2-hls-overview/}
\end{botherref}
\endbibitem

%%% 44
\bibitem[\protect\citeauthoryear{Griffiths et~al.}{2020}]{griffiths2020towards}
\begin{barticle}
\bauthor{\bsnm{Griffiths}, \binits{P.}},
\bauthor{\bsnm{Nendel}, \binits{C.}},
\bauthor{\bsnm{Pickert}, \binits{J.}},
\bauthor{\bsnm{Hostert}, \binits{P.}}:
\batitle{{Towards National-Scale Characterization of Grassland Use Intensity
  From Integrated Sentinel-2 and Landsat Time Series}}.
\bjtitle{{Remote Sensing of Environment}}
\bvolume{238},
\bfpage{111124}
(\byear{2020})
\end{barticle}
\endbibitem

%%% 45
\bibitem[\protect\citeauthoryear{Tulbure et~al.}{2022}]{tulbure2022can}
\begin{barticle}
\bauthor{\bsnm{Tulbure}, \binits{M.G.}},
\bauthor{\bsnm{Broich}, \binits{M.}},
\bauthor{\bsnm{Perin}, \binits{V.}},
\bauthor{\bsnm{Gaines}, \binits{M.}},
\bauthor{\bsnm{Ju}, \binits{J.}},
\bauthor{\bsnm{Stehman}, \binits{S.V.}},
\bauthor{\bsnm{Pavelsky}, \binits{T.}},
\bauthor{\bsnm{Masek}, \binits{J.G.}},
\bauthor{\bsnm{Yin}, \binits{S.}},
\bauthor{\bsnm{Mai}, \binits{J.}}, \betal:
\batitle{{Can We Detect More Ephemeral Floods With Higher Density Harmonized
  Landsat Sentinel 2 Data Compared to Landsat 8 Alone?}}
\bjtitle{{ISPRS Journal of Photogrammetry and Remote Sensing}}
\bvolume{185},
\bfpage{232}--\blpage{246}
(\byear{2022})
\end{barticle}
\endbibitem

%%% 46
\bibitem[\protect\citeauthoryear{Kearney et~al.}{2022}]{kearney2022monitoring}
\begin{barticle}
\bauthor{\bsnm{Kearney}, \binits{S.P.}},
\bauthor{\bsnm{Porensky}, \binits{L.M.}},
\bauthor{\bsnm{Augustine}, \binits{D.J.}},
\bauthor{\bsnm{Gaffney}, \binits{R.}},
\bauthor{\bsnm{Derner}, \binits{J.D.}}:
\batitle{{Monitoring Standing Herbaceous Biomass and Thresholds in Semiarid
  Rangelands From Harmonized Landsat 8 and Sentinel-2 Imagery to Support
  Within-Season Adaptive Management}}.
\bjtitle{{Remote Sensing of Environment}}
\bvolume{271},
\bfpage{112907}
(\byear{2022})
\end{barticle}
\endbibitem

%%% 47
\bibitem[\protect\citeauthoryear{Zhou et~al.}{2019}]{zhou2019monitoring}
\begin{barticle}
\bauthor{\bsnm{Zhou}, \binits{Q.}},
\bauthor{\bsnm{Rover}, \binits{J.}},
\bauthor{\bsnm{Brown}, \binits{J.}},
\bauthor{\bsnm{Worstell}, \binits{B.}},
\bauthor{\bsnm{Howard}, \binits{D.}},
\bauthor{\bsnm{Wu}, \binits{Z.}},
\bauthor{\bsnm{Gallant}, \binits{A.L.}},
\bauthor{\bsnm{Rundquist}, \binits{B.}},
\bauthor{\bsnm{Burke}, \binits{M.}}:
\batitle{{Monitoring Landscape Dynamics in Central Us Grasslands With
  Harmonized Landsat-8 and Sentinel-2 Time Series Data}}.
\bjtitle{{Remote Sensing}}
\bvolume{11}(\bissue{3}),
\bfpage{328}
(\byear{2019})
\end{barticle}
\endbibitem

%%% 48
\bibitem[\protect\citeauthoryear{Zhu and Ye}{}]{physnews}
\begin{botherref}
\oauthor{\bsnm{Zhu}, \binits{Z.}},
\oauthor{\bsnm{Ye}, \binits{S.}}:
{AI satellite mapping can quickly pinpoint hurricane damage across an entire
  state to spot where people may be trapped}.
\url{https://phys.org/news/2022-10-ai-satellite-quickly-hurricane-entire.html}
\end{botherref}
\endbibitem

%%% 49
\bibitem[\protect\citeauthoryear{Freitag et~al.}{2022}]{freitag2022efficient}
\begin{botherref}
\oauthor{\bsnm{Freitag}, \binits{M.O.}},
\oauthor{\bsnm{Albrecht}, \binits{C.M.}},
\oauthor{\bsnm{Marianno}, \binits{F.J.}},
\oauthor{\bsnm{Lu}, \binits{S.}},
\oauthor{\bsnm{Hamann}, \binits{H.F.}},
\oauthor{\bsnm{Schmude}, \binits{J.W.}}:
{Efficient Querying Using Overview Layers of Geospatial-Temporal Data in a Data
  Analytics Platform}.
Google Patents.
US Patent 11,360,970
(2022)
\end{botherref}
\endbibitem

%%% 50
\bibitem[\protect\citeauthoryear{PRISM Climate~Group}{}]{PRISM}
\begin{botherref}
\oauthor{\bsnm{PRISM Climate~Group}, \binits{O.S.U.}}:
{Prism}.
\url{https://prism.oregonstate.edu}
\end{botherref}
\endbibitem

%%% 51
\bibitem[\protect\citeauthoryear{Claverie
  et~al.}{2017}]{claverie2017harmonized}
\begin{botherref}
\oauthor{\bsnm{Claverie}, \binits{M.}},
\oauthor{\bsnm{Masek}, \binits{J.G.}},
\oauthor{\bsnm{Ju}, \binits{J.}},
\oauthor{\bsnm{Dungan}, \binits{J.L.}}:
{Harmonized Landsat-8 Sentinel-2 (HLS) Product User’s Guide}.
{National Aeronautics and Space Administration (NASA): Washington, DC, USA}
(2017)
\end{botherref}
\endbibitem

%%% 52
\bibitem[\protect\citeauthoryear{He et~al.}{2022}]{mae}
\begin{bchapter}
\bauthor{\bsnm{He}, \binits{K.}},
\bauthor{\bsnm{Chen}, \binits{X.}},
\bauthor{\bsnm{Xie}, \binits{S.}},
\bauthor{\bsnm{Li}, \binits{Y.}},
\bauthor{\bsnm{Doll{\'a}r}, \binits{P.}},
\bauthor{\bsnm{Girshick}, \binits{R.}}:
\bctitle{{Masked Autoencoders Are Scalable Vision Learners}}.
In: \bbtitle{Proceedings of the IEEE/CVF Conference on Computer Vision and
  Pattern Recognition},
pp. \bfpage{16000}--\blpage{16009}
(\byear{2022})
\end{bchapter}
\endbibitem

%%% 53
\bibitem[\protect\citeauthoryear{Tong et~al.}{2022}]{videomae}
\begin{botherref}
\oauthor{\bsnm{Tong}, \binits{Z.}},
\oauthor{\bsnm{Song}, \binits{Y.}},
\oauthor{\bsnm{Wang}, \binits{J.}},
\oauthor{\bsnm{Wang}, \binits{L.}}:
{Videomae: Masked Autoencoders Are Data-Efficient Learners for Self-Supervised
  Video Pre-Training}.
{Preprint Available on arXiv:2203.12602}
(2022)
\end{botherref}
\endbibitem

%%% 54
\bibitem[\protect\citeauthoryear{Feichtenhofer
  et~al.}{2022}]{spatiotemporalmae}
\begin{bchapter}
\bauthor{\bsnm{Feichtenhofer}, \binits{C.}},
\bauthor{\bsnm{fan}, \binits{h.}},
\bauthor{\bsnm{Li}, \binits{Y.}},
\bauthor{\bsnm{He}, \binits{K.}}:
\bctitle{{Masked Autoencoders as Spatiotemporal Learners}}.
In: \bbtitle{Advances in Neural Information Processing Systems},
vol. \bseriesno{35},
pp. \bfpage{35946}--\blpage{35958}
(\byear{2022})
\end{bchapter}
\endbibitem

%%% 55
\bibitem[\protect\citeauthoryear{Bonafilia
  et~al.}{2020}]{bonafilia2020sen1floods11}
\begin{bchapter}
\bauthor{\bsnm{Bonafilia}, \binits{D.}},
\bauthor{\bsnm{Tellman}, \binits{B.}},
\bauthor{\bsnm{Anderson}, \binits{T.}},
\bauthor{\bsnm{Issenberg}, \binits{E.}}:
\bctitle{{Sen1Floods11: A Georeferenced Dataset to Train and Test Deep Learning
  Flood Algorithms for Sentinel-1}}.
In: \bbtitle{Proceedings of the IEEE/CVF Conference on Computer Vision and
  Pattern Recognition Workshops},
pp. \bfpage{210}--\blpage{211}
(\byear{2020})
\end{bchapter}
\endbibitem

%%% 56
\bibitem[\protect\citeauthoryear{Contributors}{2020}]{mmseg2020}
\begin{botherref}
\oauthor{\bsnm{Contributors}, \binits{M.}}:
{{MMSegmentation}: OpenMMLab Semantic Segmentation Toolbox and Benchmark}.
\url{https://github.com/open-mmlab/mmsegmentation}
(2020)
\end{botherref}
\endbibitem

%%% 57
\bibitem[\protect\citeauthoryear{Isola et~al.}{2017}]{pix2pix2017}
\begin{bchapter}
\bauthor{\bsnm{Isola}, \binits{P.}},
\bauthor{\bsnm{Zhu}, \binits{J.-Y.}},
\bauthor{\bsnm{Zhou}, \binits{T.}},
\bauthor{\bsnm{Efros}, \binits{A.A.}}:
\bctitle{{Image-to-Image Translation With Conditional Adversarial Networks}}.
In: \bbtitle{Proceedings of the IEEE Conference on Computer Vision and Pattern
  Recognition},
pp. \bfpage{1125}--\blpage{1134}
(\byear{2017})
\end{bchapter}
\endbibitem

%%% 58
\bibitem[\protect\citeauthoryear{Baier et~al.}{2022}]{baier2020building}
\begin{barticle}
\bauthor{\bsnm{Baier}, \binits{G.}},
\bauthor{\bsnm{Deschemps}, \binits{A.}},
\bauthor{\bsnm{Schmitt}, \binits{M.}},
\bauthor{\bsnm{Yokoya}, \binits{N.}}:
\batitle{{Synthesizing Optical and SAR Imagery From Land Cover Maps and
  Auxiliary Raster Data}}.
\bjtitle{{IEEE Transactions on Geoscience and Remote Sensing}}
\bvolume{60},
\bfpage{1}--\blpage{12}
(\byear{2022})
\end{barticle}
\endbibitem

%%% 59
\bibitem[\protect\citeauthoryear{Heusel
  et~al.}{}]{DBLP:journals/corr/HeuselRUNKH17}
\begin{botherref}
\oauthor{\bsnm{Heusel}, \binits{M.}},
\oauthor{\bsnm{Ramsauer}, \binits{H.}},
\oauthor{\bsnm{Unterthiner}, \binits{T.}},
\oauthor{\bsnm{Nessler}, \binits{B.}},
\oauthor{\bsnm{Klambauer}, \binits{G.}},
\oauthor{\bsnm{Hochreiter}, \binits{S.}}:
{GANs Trained by a Two Time-Scale Update Rule Converge to a Nash Equilibrium}.
{Advances in Neural Information Processing Systems}
\textbf{30},
1--12
\end{botherref}
\endbibitem

%%% 60
\bibitem[\protect\citeauthoryear{Liu et~al.}{2021}]{liu2021swin}
\begin{bchapter}
\bauthor{\bsnm{Liu}, \binits{Z.}},
\bauthor{\bsnm{Lin}, \binits{Y.}},
\bauthor{\bsnm{Cao}, \binits{Y.}},
\bauthor{\bsnm{Hu}, \binits{H.}},
\bauthor{\bsnm{Wei}, \binits{Y.}},
\bauthor{\bsnm{Zhang}, \binits{Z.}},
\bauthor{\bsnm{Lin}, \binits{S.}},
\bauthor{\bsnm{Guo}, \binits{B.}}:
\bctitle{{Swin Transformer: Hierarchical Vision Transformer Using Shifted
  Windows}}.
In: \bbtitle{Proceedings of the IEEE/CVF International Conference on Computer
  Vision},
pp. \bfpage{10012}--\blpage{10022}
(\byear{2021})
\end{bchapter}
\endbibitem

%%% 61
\bibitem[\protect\citeauthoryear{Chen et~al.}{2017}]{chen2017rethinking}
\begin{botherref}
\oauthor{\bsnm{Chen}, \binits{L.-C.}},
\oauthor{\bsnm{Papandreou}, \binits{G.}},
\oauthor{\bsnm{Schroff}, \binits{F.}},
\oauthor{\bsnm{Adam}, \binits{H.}}:
{Rethinking Atrous Convolution for Semantic Image Segmentation}.
{Preprint Available on arXiv:1706.05587}
(2017)
\end{botherref}
\endbibitem

%%% 62
\bibitem[\protect\citeauthoryear{Ronneberger
  et~al.}{2015}]{10.1007/978-3-319-24574-4_28}
\begin{bchapter}
\bauthor{\bsnm{Ronneberger}, \binits{O.}},
\bauthor{\bsnm{Fischer}, \binits{P.}},
\bauthor{\bsnm{Brox}, \binits{T.}}:
\bctitle{U-net: Convolutional networks for biomedical image segmentation}.
In: \bbtitle{Proceedings of the 18th International Conference on Medical Image
  Computing and Computer-Assisted Intervention (MICCAI)},
pp. \bfpage{234}--\blpage{241}
(\byear{2015})
\end{bchapter}
\endbibitem

%%% 63
\bibitem[\protect\citeauthoryear{Li et~al.}{2023}]{li2023assessment}
\begin{botherref}
\oauthor{\bsnm{Li}, \binits{W.}},
\oauthor{\bsnm{Lee}, \binits{H.}},
\oauthor{\bsnm{Wang}, \binits{S.}},
\oauthor{\bsnm{Hsu}, \binits{C.-Y.}},
\oauthor{\bsnm{Arundel}, \binits{S.T.}}:
{Assessment of IBM and NASA's Geospatial Foundation Model in Flood Inundation
  Mapping}.
{Preprint Available on arXiv:2309.14500}
(2023)
\end{botherref}
\endbibitem

\end{thebibliography}

\clearpage

\begin{appendices}

\end{appendices}

\end{document}